\documentclass{article}

     \usepackage[final, nonatbib]{neurips_2021}

\usepackage[utf8]{inputenc} % allow utf-8 input
\usepackage[T1]{fontenc}    % use 8-bit T1 fonts
\usepackage{hyperref}       % hyperlinks
\usepackage{url}            % simple URL typesetting
\usepackage{booktabs}       % professional-quality tables
\usepackage{amsfonts}       % blackboard math symbols
\usepackage{nicefrac}       % compact symbols for 1/2, etc.
\usepackage{microtype}      % microtypography
\usepackage{xcolor}         % colors

\usepackage{graphicx}
\usepackage{amsthm}
\usepackage{amsmath}

\usepackage{array}
\usepackage{makecell}
\usepackage{multicol, blindtext}
\usepackage{soul,color,xcolor}
\usepackage{multirow}
\usepackage[english]{babel}
\usepackage{amssymb}
\usepackage{tabularx}
\usepackage{bm}
\usepackage{comment}
\usepackage{float}
\usepackage{algorithm}
\usepackage{caption}
\usepackage{wrapfig}
\usepackage{cite}
\usepackage{adjustbox}
\usepackage{bbm}
\usepackage{subfig}

\hypersetup{
    colorlinks=true,
    citecolor = blue,
    linkcolor=blue
}

\usepackage{mathtools, nccmath}
\usepackage{algorithm,algorithmicx,algpseudocode}

\usepackage{amsthm}

\newtheorem{definition}{Definition}[section]
\newtheorem{assumption}{Assumption}[section]

\newcommand{\squeeze}[1]{{#1\parfillskip=0pt\par}}

\title{Invariant Causal Imitation Learning\\for Generalizable Policies}

\author{%
        Ioana Bica\thanks{Equal contribution.} \\
        University of Oxford, Oxford, UK\\
        The Alan Turing Institute, London, UK \\
        \texttt{ioana.bica@eng.ox.ac.uk} \\
        \And
        Daniel Jarrett$^*$ \\
        University of Cambridge, Cambridge, UK\\
        \texttt{daniel.jarrett@maths.cam.ac.uk} \\
        \And
        Mihaela van der Schaar \\
        University of Cambridge, Cambridge, UK\\
        University of California, Los Angeles, USA \\
        The Alan Turing Institute, London, UK \\
        \texttt{mv472@cam.ac.uk} \\
}

\begin{document}

\maketitle

\begin{abstract}
\squeeze{Consider learning an imitation policy on the basis of demonstrated behavior from multiple environments, with an eye towards deployment in an unseen environment. Since the observable features from each setting may be different, directly learning individual policies as mappings from features to actions is prone to \textit{spurious correlations}---and may not generalize well. However, the expert’s policy is often a function of a shared \textit{latent structure} underlying those observable features that is invariant across settings.
By leveraging data from multiple environments, we propose \textit{Invariant Causal Imitation Learning} (ICIL), a novel technique in which we learn a feature representation that is invariant across domains, on the basis of which we learn an imitation policy that matches expert behavior.
To cope with transition dynamics mismatch, ICIL learns a \textit{shared} representation of causal features (for all training environments), that is independent from the \textit{specific} representations of noise variables (for each of those environments).
Moreover, to ensure that the learned policy matches the observation distribution of the expert's policy, ICIL estimates the energy of the expert's observations and uses a regularization term that minimizes the imitator policy's next state energy.
Experimentally, we compare our methods against several benchmarks in control and healthcare tasks and show its effectiveness in learning imitation policies capable of generalizing to unseen environments.}
    % Consider the problem of learning an imitation policy in the strictly batch setting that is capable of generalizing across environments. 
    % In many cases, the expert's policy depends on a shared latent structure of the observations that causally influence the expert's actions. This causal structure is shared and remains invariant across environments. However, demonstrated trajectories from an expert's policy in a single environment can have spurious correlations and biases. By leveraging data from multiple environments, we propose Invariant Causal Imitation Learning (ICIL), a new methods that learn a representation of observations that is invariant across domains and an imitation policy that depends on this causal representation and that matches the demonstrator's behaviour. To allow for transition dynamics mismatch, ICIL learns a shared causal representation and different disentangled representations of the spurious correlations across the training environments. Moreover, to ensure that the learnt imitation policy matches the observation distribution of the expert's policy, ICIL estimates the energy of the expert's observations and uses a regularization term that minimizes the imitator policy's next state energy. Experimentally, we compare our methods against several benchmarks in control and healthcare tasks and show its effectiveness in learning imitation policies that can generalizing to unseen environments. 
\end{abstract}

\section{Introduction}

Strictly batch imitation learning aims to learn a policy that directly mimics the behaviour of experts, for which we only have access to a set of demonstrations: logged trajectories of observations and actions following the expert's policy \cite{piot2016bridging, kostrikov2019imitation, jarrett2020strictly}. We cannot interact online with the environment, let alone query the expert any further, nor do we have reward signals for supervision. This setting is relevant in real-world scenarios where live experimentation is risky or costly---such as healthcare and education.

\squeeze{Our aim is to learn an imitation policy in the strictly batch setting that faithfully matches the expert behaviour, while at the same time is able to generalize to unseen environments. In healthcare, learning a generalizable behaviour policy that could achieve expert performance in new environments is an important goal: As a means of providing clinical decision support, it could serve as an ``individualized'' clinical guideline for actions that can be taken for different patients---especially in a hospital, region, or patient demographic from which we have no access to data during training. In this endeavor, a principal challenge is that the sets of expert demonstrations that we have access to may contain variables that induce selection bias, or are otherwise spuriously correlated with the expert's actions \cite{zhang2018natural, torralba2011unbiased, arjovsky2019invariant, zhang2020invariant}. Directly learning an imitation policy from such data may lead to learning those spurious associations, thereby failing to generalize to unseen environments, and perpetuating any biases in the expert's behaviour.}

However, in general it is likely that the expert's actions are only causally affected by a subset of the observed variables or by a shared latent structure \cite{de2019causal, lee2021causal}. For instance, when imitating ideal driving behaviour, the background scenery might change, but the actions should only depend on car and road features. Another example includes the case when the lightning conditions in a room are changing, but physical dynamics of the environment are staying the same 
\cite{zhang2020invariant}. By leveraging expert trajectories from multiple different environments, our aim is to uncover this shared latent structure that causally determines expert actions, which allows us to eliminate the spurious associations and biases. In this way, the learnt policy will better be able to generalize to any unseen environments that share the same latent structure as those used for training.

\begin{figure}%
    \centering
    \includegraphics[width=0.70\textwidth]{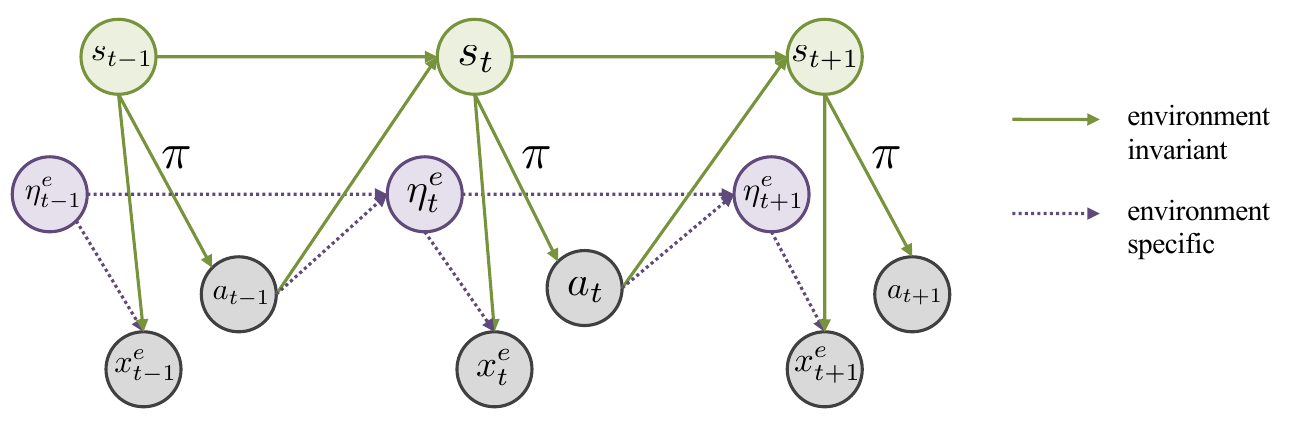}%
    \caption{\squeeze{Causal diagram for the structure of environments. Expert demonstrations contain information about observations $x_t$ and actions $a_t$. We assume that observations are decomposable into (1) state representations $s_t$ that consist of the causal parents of the actions, and (2) noise representations $\eta_t$ that encapsulate any spurious correlations with the actions. To allow for dynamics mismatch, the transitions between the noise representations are specific to each environment. We want to recover the invariant state representation $s_t$ such that the learned policy $\pi(\cdot|s_t)$ generalizes well to new environments.}}
    \label{fig:causal_diagram}
    \vspace{-8mm}
\end{figure}

As illustrated in Figure \ref{fig:causal_diagram}, we assume access to observations and actions from the expert's policy in the different environments $e$. The observations are functions of noise factors $\eta^e$ (which may differ across environments) and shared latent state representations $s$ (which is invariant across environments)---that encapsulate the causal parents of the expert's actions. Note that the observed features for an environment may simply be the union of $\eta^e$ and $s$, but they may also be any non-linear transformation of them. We shall operate in the setting where there are no hidden confounders, i.e. that we observe all variables that are affecting the expert's actions (and the next states that result from these actions).

In addition to spurious correlations, another difficulty stems from learning to imitate sequential behavior in the strictly batch setting itself: While behaviour cloning \cite{pomerleau1991efficient} provides an intrinsically batch solution, it ignores important information contained in the expert's roll-out distribution, and the learned policy may drift from the support of the distribution of states visited by the expert \cite{ross2010efficient, ross2011reduction}.

\squeeze{\textbf{Contributions:} In this paper, we introduce \textit{Invariant Causal Imitation Learning} (ICIL), a novel method that learns a causal representation of the expert's actions---which is used to build a generalizable imitation policy that matches the expert's behaviour. ICIL operates in the strictly batch setting and does not assume access to data from the target environments. By leveraging expert demonstrations from multiple different training environments, ICIL learns an (shared) invariant causal representation as well as an (environment-specific) noise representation. This accommodates dynamics mismatch across environments, while allowing the imitation policy to be learned by conditioning on the invariant causal representations. First, to satisfy the causal relationships in Figure \ref{fig:causal_diagram}, ICIL learns dynamics preserving representations and ensures that the learnt causal and noise representations are marginally independent by minimizing their mutual information. Second, to encourage the learnt imitation policy to stay within the support of the distribution of states visited by the expert's policy, ICIL estimates the energy of the expert’s observations and uses a regularization term that minimizes the imitator policy’s next state energy. Third, we evaluate ICIL against benchmarks for batch imitation learning in control and healthcare environments. We also empirically investigate directly using ideas from invariant risk minimization \cite{arjovsky2019invariant} to augment the loss function of existing batch imitation learning methods, and benchmark against their ability to generalize across environments.}

%Make more formal the set-up: vairable that affect the action and variables that we observe in the action. Causal features is a subset of all variable. Specify that we have no hidden confounders. 

%Emphasize that its not separate variables, but also a function. 

%Make it clear what the two big challenges are (1) spurious correlations and (2) batch setting. 

%Make the related works about these challenges: start with related work on invariant risk minimization (they don't do imitation), literature on batch imitation learning (describe limitation - do not take into account spurious correlations). Then say that works in domain adaptation/transfer in imitation and causality in imitation. 

%Related works - say they assume access to reward, assume access to interactions, assume access to interactions. 

%Possible extensions: 
%\begin{itemize}
%    \item build causal representation of entire history
%    \item consider hidden confounders (like in the paper of Alexis). This will require a POMDP setting. 
%\end{itemize}

\section{Related Works}

%Put table 1 after the formalism to make it more technical. 

%Very clear understanding of what is assumed and what is learnt? Are our assumptions similar to other people's assumptions. 

%An important piece to make this interesting is across environments and over time. 

%Highlight other works that do IRM to emphasize the temporal component.

%The main differences/characteristics in our set-up:
%\begin{itemize}
%    \item We consider the batch setting: no access to a model, no ability to simulate extra trajectories.
%    \item We only have access to trajectories from a limited set of environments (domains) and we want our method to work in other environments (domains) from which we do not have access to data. 
%    \item We assume a certain structure to the environments (domains) and the data that we have from these environments. In particular, we assume that there is some latent structure shared across all environments. %   \item We allow for dynamics mismatch between environments (from the noise variables).
%    \item We match expert state distribution.  
%\end{itemize}

We tackle the problem of learning generalizable policies in an offline setting using ideas from causal inference. As such, our work straddles the intersection of research in (1) strictly batch imitation, (2) invariant representation learning, and---more broadly---(3) causality in sequential decision-making.

\textbf{Strictly Batch Imitation Learning}: The simplest approach to imitation learning in the batch setting is behaviour cloning \cite{pomerleau1991efficient} which uses standard supervised learning techniques to learn an imitation policy that minimizes  the  negative  log-likelihood  of  the  observed  demonstrator  actions. However, behaviour cloning suffers from distributional shift as the learnt imitation policy cannot recover if it reaches a state out-of the distribution of the expert demonstrations \cite{melo2010learning, ross2010efficient, ross2011reduction, piot2014boosted}. To overcome this problem, \cite{piot2014boosted, piot2016bridging} propose incorporate dynamics-awareness by adding regularization to behaviour cloning by using norm-based penalties on the sparsity of implied rewards. Alternatively, \cite{kostrikov2019imitation} uses a distribution matching approach and propose an offline objective for estimating the distribution ratio of the imitator policy and the expert policy, while \cite{jarrett2020strictly} jointly learn a policy function together with an energy-based model of the state distribution. However, none of the existing approaches consider the problem of \textit{generalization} across environments and learning policies robust to spurious correlations. 

% Change to "Access to target trajectories"

\newcolumntype{A}{>{\centering\arraybackslash}m{0.9 cm}}
\begin{table*}[t]
\begin{center}
\begin{small}
\setlength\tabcolsep{1pt}
\begin{adjustbox}{max width=\textwidth}
\begin{tabular}{Alccccccc} 
\toprule
& Method & Environment & Offline & \makecell{Dynamics\\ mismatch} & \makecell{Sensory-shift \\ (hidden confounders)} & \makecell{State-distribution\\ matching} & \makecell{No access to\\ target trajectories}  & \makecell{Temporal\\ aspect} \\
\midrule
\multirow{4}{*}{\rotatebox[origin=c]{90}{\scalebox{0.8}{\makecell{Imitation\\Learning}}}}
& Pomerleau \cite{pomerleau1991efficient} & Model-free & Yes & No & No & No & N/A & No \\
& Ho \& Ermon \cite{ho2016generative} & Model-free & No & No & No & Model rollouts & N/A & Yes \\
& Kostrikov et al. \cite{kostrikov2019imitation} & Model-free & Yes & No & No & Adversarial off-policy matching & N/A & Yes \\
& de Haan \textit{et al.} \cite{de2019causal} & Model-free & No & No & No & No & N/A & Yes \\
\midrule
\multirow{3}{*}{\rotatebox[origin=c]{90}{\scalebox{0.8}{\makecell{Generaliz.\\in IL}}}}
& Lu \textit{et al.} \cite{lu2020adail}  & Model-free  & No & Yes & No & Model rollouts & No & Yes \\
& Kim \textit{et al.} \cite{kim2020domain} & Model-based & No & Yes & No  & Model rollouts & No & Yes  \\
& Etsami \textit{et al.} \cite{etesami2020causal} & Model-free & Yes & No &  Yes & No & No & Yes \\
\midrule
IRM
& Arjovsky \textit{et al.}\cite{arjovsky2019invariant} & Model-free & Yes & N/A & N/A & N/A & Yes & No \\
\midrule
 \multicolumn{2}{l}{~~ICIL (Ours)} & Model-based & \textbf{Yes} &  \textbf{Yes}  &   No & Energy-based & \textbf{Yes} & \textbf{Yes} \\
\bottomrule
\end{tabular}
\end{adjustbox}
\end{small}
\end{center}
\vspace{-0.75em}
\caption{Comparison of our proposed method with related works. ICIL operates in the strictly batch setting, allows for dynamics mismatch, does not require access to target trajectories, and incentivizes the imitation policy to stay in the support of the expert's distribution via energy-based regularization.}
\label{tab:related-works}
\vspace{-2mm}
\end{table*}

\squeeze{\textbf{Invariant Risk Minimization}: In the supervised learning setting, Invariant Risk Minimization (IRM) \cite{arjovsky2019invariant} leverages data from multiple domains to learn a data representation that elicits an invariant predictor across the different environments. The training data from each environment corresponds to different interventions on the data generating process. Given data from several training environments, the IRM objective aims to find a representation such that there exists a classifier that is optimal across all training domains, i.e. that minimizes the empirical risk in each domain.  This represents a challenging, bi-level optimization, and \cite{arjovsky2019invariant} propose the IRM-v1 objective which is a practical version to optimize. Through this optimization, the IRM objective should learn a predictor that only uses the causal parents of the target variable and that is thus invariant across environments. However, directly using IRM for our \textit{sequential} problem setting is not desirable, since it does not take into account the effect of each action on the subsequent states. Nonetheless, we empirically investigate augmenting existing methods for batch imitation to use the IRM-v1 objective in conjunction with their defined imitation risk, and verify whether they are able to generalize across environments. In our experiments we observe that, in general, directly applying the IRM objective in this manner is not good enough.}

\textbf{Generalization in Imitation Learning:} The problem of domain adaptation and transfer learning for the imitation learning setting has been tackled by several works so far. However, while they consider problems of dynamics-, embodiment-, and/or viewpoint-mismatch between the imitator and expert, existing methods assume access to demonstrations from the target environment \cite{kim2020domain, sermanet2017time, liu2018imitation}, assume access to online interaction or simulators in the different environments \cite{lu2020adail}, or focus on the different problem of hidden confounding \cite{etesami2020causal, zhang2020causal}. Another line of work that is related is learning from demonstrations and meta-learning. While works in meta-learning also aims to generalize learnt policies to new-tasks, they require access to one or more expert trajectories from the new task
% , in zero-shot or few shot learning
\cite{finn2017model, duan2017one, finn2017one, yu2018one, james2018task, sharma2019third}.

%Meta-Learning for Domain Generalization \cite{li2018learning}: trains a base learner on a set of source domains by synthesizing virtual training and virtual testing domains within each mini-batch. Code: https://github.com/HAHA-DL/MLDG

\squeeze{\textbf{Causality in Imitation Learning and Reinforcement Learning:} Several ideas from causality have been used to improve imitation learning and generalization in reinforcement learning. The idea of conditioning the imitation policy on the causal parents has been employed by \cite{de2019causal} to avoid the problem of `causal confusion' when learning a policy for the single environment setting. However, \cite{de2019causal} requires queering the expert or being able to perform interventions in the environment this is not possible in the batch setting. Similarly, \cite{lee2021causal, volodin2020resolving} also learn causal relationships between the observations, actions and rewards by performing/simulating the effect of interventions in the environment. Alternatively, \cite{sonar2020invariant} use ideas from Invariant Risk Minimization \cite{arjovsky2019invariant, ahuja2020invariant} to learn optimal reinforcement learning policies that generalize across domains. Perhaps the most similar setting to ours is the one in \cite{zhang2020invariant} which studies the problem of generalization in reinforcement learning and also learn a representation that is shared across the domains. However, unlike our imitation setting, they assume access to a \textit{known} reward signal, and focus on learning the causal ancestors of that reward to improve reinforcement learning \cite{haarnoja2018soft}.}

%\begin{itemize}
%\item Invariant Causal Prediction for Block MDPs \cite{zhang2020invariant}
%\item Learning Invariant Representations for Reinforcement Learning without Reconstruction \cite{zhang2020learning}
%\item Invariant Policy Optimization: Towards Stronger Generalization in Reinforcement Learning \cite{sonar2020invariant}
%\item Resolving Spurious Correlations in
%Causal Models of Environments via Interventions \cite{volodin2020resolving}
%\item Causal Reasoning in Simulation for Structure and Transfer Learning of Robot Manipulation Policies \cite{lee2021causal}
%\end{itemize}

To the best of our knowledge, we are the first to tackle the problem of learning generalizable imitation policies in the strictly batch setting. Table \ref{tab:related-works} summarizes main differences with relevant related works.
% Additional related works can be found in Appendx \ref{apx:related_works}. {\color{red} Dan, do you think we need to have other related works in the appendix?}

%To summarize, ICIL  works in the strictly batch setting imitation learning where we have no access to the reward, a simulator model nor the ability to simulate extra trajectories. Moreover, we assume access to demonstrations from a limited set of environments and our method is capable of generalizing to other environments (domains) from which we do not have access to data. In addition, we allow for dynamics mismatch between the different environments we learn from and generalize to. Finally, to enable imitation learning and matching the behaviour of the expert, we incorporate an objective that aims to match the expert's observations distribution. 

%Directly applying IRM involves adding the IRM penalty to behaviour cloning. 

%What we observe that is different, what needs to be matched. Consider what needs to be optimized. How is causality helpful in our frameworks. 

%Consider the cases when we can transfer. 
 
\section{Problem Formalism}

\subsection{Imitation Learning}

We work in the standard Markov decision process (MDP) setting: Let an environment be given by $e=(\mathcal{X}, \mathcal{A}, T, r, \gamma)$, with observations $x\in \mathcal{X}$, actions $a\in \mathcal{A}$, transition function $T \in \Delta(\mathcal{X})^{\mathcal{X}\times\mathcal{A}}$, reward function $r \in \mathbb{R}^{\mathcal{X}\times\mathcal{A}}$, and discount factor $\gamma$. Let $\pi\in\Delta(\mathcal{A})^{\mathcal{X}}$ be a policy with the induced occupancy measure $\rho_{\pi}(x, a) =(1-\gamma)\sum_{t=0}^{\infty}\gamma^{t}p(x_{t}=x,a_{t}=a|x_{t}$$\sim$$ T(\cdot|x_{t-1},a_{t-1}),a_{t}$$\sim$$\pi(\cdot|x_{t}))$ of observations and actions, and let $\rho_{\pi}(x) = \sum_{a\in \mathcal{A}} \rho_{\pi}(x, a)$ be the observation occupancy measure.

\squeeze{Unlike in the reinforcement learning setting, where the aim is to learn a policy $\pi(\cdot \mid x)$ that maximizes the cumulative sum of some known reward signal, in imitation learning the reward is neither known nor observed. Instead, we only have access to a dataset of trajectories $\mathcal{D} = \{\tau_{i}\}_{i=1}^{N}$ from a demonstrator policy $\pi_D$, where each trajectory $\tau \sim \pi_D = (x_t, a_t, x_{t+1})_{t=0, \dots}$ consists of a sequence of observation, action, next observation tuples that are sampled as $a_t \sim \pi_D (\cdot \mid x_t)$ and $x_{t+1} \sim T(\cdot \mid x_t, a_t)$.} 

The goal of imitation learning is to seek an imitation policy $\pi$ that minimizes the following risk:
\begin{equation}
    R(\pi) = \mathcal{L}(\pi, \pi_D)
\end{equation}
where $\mathcal{L}$ is a choice of loss function. Now, if we were in the online setting, we would have access to the environment (or a simulator), with which we can interactively perform distribution matching by minimizing the divergence between the expert's state occupancy $\rho_D$ and the imitator's state occupancy $\rho_\pi$ \cite{ho2016generative, baram2016model, finn2016connection, ghasemipour2020divergence}. One example is to use the (forward) KL divergence: $\mathcal{L}(\pi, \pi_D) = D_{KL}(\rho_D || \rho_{\pi})$ \cite{ghasemipour2020divergence}. However, in the offline setting we have no further access to the environment. As noted above, the simplest solution is behaviour cloning (BC) \cite{pomerleau1991efficient, bain1995framework, syed2010reduction}, which minimizes the negative log-likelihood of the demonstrator's actions. However, by disregarding the distribution of the expert's observations, imitation policies learnt by BC often result in compounding error when deployed in practice \cite{melo2010learning, ross2010efficient, ross2011reduction, piot2014boosted}.

%+ give some examples here for what this loss can be (gives examples as BC loss and )

%In online setting with expert is DAgger, without expert distribution matching. In offline the simplest thing is to do BC and then problems we solve later. 

\subsection{Imitation Learning from Multiple Environments}

\squeeze{Consider a \textit{family} of environments $\mathcal{M} = \{(\mathcal{X}^e, \mathcal{A}, T^e, r^e, \gamma) \mid e \in \mathcal{E} \}$ with observations $x^{e} \in \mathcal{X}^e$, actions $a \in \mathcal{A}$, transition function $T^{e} \in \Delta(\mathcal{X})^{\mathcal{X}\times\mathcal{A}}$, reward function $r^{e} \in \mathbb{R}^{\mathcal{X}\times\mathcal{A}}$, and discount factor $\gamma$. This is the primary setting that we shall operate in. Note that the action space and discount factor do not change between environments. For notational simplicity, when considering the union over environments, we shall drop the index $e$. 
%Let $\pi: \mathcal{X} \times \mathcal{A} \rightarrow $ be a policy that induces the following occupancy measure for the observed variables in environment $e$: $\rho^e_{\pi}(x, a) = \mathbb{E}_{\pi} [\sum_{t=0}^{\infty} \gamma^t \mathbbm{1}_{\{x^e_t=x, a_t=a\}}]$ where the expectation is taken over $a_t \sim \pi(\cdot \mid x_t)$ and $x^e_{t+1} \sim T^e(\cdot \mid x^e_t, a_t)$. Let $\pi_E$ be the expert policy. 
We assume offline access to a dataset of recorded trajectories from the expert policy $\pi_D$ in a set of training environments $\mathcal{E}_{train}$$\subset$$\mathcal{E}$, $\mathcal{D} = \{ \{(\tau^{e}_{i}\}_{i=1}^{N_e} \mid e \in \mathcal{E}_{train} \}$. Each trajectory $\tau^e \sim \pi_D = (x^{e}_t, a_t, x^{e}_{t+1})_{t=0,\dots}$ consists of a sequence of environment specific observations, expert actions and next observations sampled as $a_t \sim \pi_D(\cdot \mid x^{e}_{t})$ and $x_{t+1}^e \sim T^{e}(\cdot \mid x^{e}_{t}, a_{t})$.}

\squeeze{In the presence of multiple environments, our goal is to learn a policy $\pi\in\Delta(\mathcal{A})^{\mathcal{X}}$ that matches the expert behaviour in all possible environments $\mathcal{E}$ that share a certain structure for the observations and the transition dynamics. In particular, this involves finding a policy that \textit{generalizes} well across these related environments $e \in \mathcal{E}$---that is, the policy should ideally minimize the imitation risk across them:}
\begin{equation} \label{eq:imitation_risk}
    \max_{e\in \mathcal{E}} R^{e}(\pi) = \mathcal{L}^{e}(\pi, \pi_D)
\end{equation}
where each $\mathcal{L}^e$ explicitly depends on the characteristics of the environment $e$. Note that since we know nothing specific about $\mathcal{E}$, it is difficult to optimize for this directly. That said, if we make mild assumptions about the ``relatedness'' of these environments, we can learn policies that generalize well.

%Note that this set-up is similar to the one in \cite{zhang2020invariant}. However, one of the main differences is that we do not observe the reward function.

\squeeze{\textbf{Structure of Observations:} First, we assume there is a shared latent structure underlying the observations from different environments---on which the expert policy depends. Finding such a structure would let us discard irrelevant factors as inputs to the learnt policy, improving generalization \cite{peters2016causal, arjovsky2019invariant, zhang2020invariant}:}

\begin{assumption}\label{assumption:latent}
\squeeze{\textit{(Shared Latent Structure)}~
Consider decomposing the observations $x^{e}
\in \mathcal{X}^e$  in each environment $e \in \mathcal{E}$ into two components: an invariant representation $s\in \mathcal{S}$ and noise terms $\eta^e \in \mathcal{Z}^e$ (i.e. spurious correlations), such that $x^{e} = q(s, \eta^e)$ for some invertible transformation $q: \mathcal{S} \times \mathcal{Z}^e \rightarrow \mathcal{X}^{e}$. There exists some $q$ such that $\pi_D$ only depends on $s$, and space $\mathcal{S}$ is non-empty.}
\end{assumption}

%In particular, we consider that the observations  in each environment $e$ can be decomposed into state variables $s\in \mathcal{S}$ and noise terms $\eta^e \in \mathcal{Z}^e$ such that $x = q^e(s, \eta^e)$, where $q^{e}: \mathcal{S} \times \mathcal{Z}^e \rightarrow \mathcal{X}$. 

\squeeze{In other words, we assume that the demonstrator's policy $\pi_{D}$ depends only on information that is shared across the environments, i.e. the state variables $s$ are the causal parents of the expert action $a \sim \pi_D (\cdot \mid s)$. As illustrated in Figure \ref{fig:causal_diagram}, the state variables and the noise terms are responsible for generating the patient observations, but the policy depends only on the state variables. Thus, we allow different environments to have different $p(x)$ marginals (as well as different $p(a|x)$). This allows environments to have different structure. The only requirement is that the environments are the same as far as the task is concerned. This means that there exists some $\mathcal{S}$ such that $p(s)$ marginals should be the same (as well as $p(a|s)$). Learning such a representation that is invariant satisfies the standard Environment Invariance Constraint \cite{creager2021environment}. While this set-up is similar to the one in \cite{zhang2020invariant}, a crucial difference is that we have no access to any reward functions whatsoever, and that we must learn an imitation policy in a strictly batch setting.}

\squeeze{Note that the latent structure induced by the state variables is \textit{shared} across the different environments. This means that the transition dynamics for the state representation $p(s_{t+1}\mid s_{t}, a_{t})$ remain invariant across the environments. On the other hand, as different environments may be characterized by different types of spurious correlations, to allow for flexibility in their structure and evolution, we consider that the transition dynamics of the noise terms $p^{e}(\eta^{e}_{t+1} \mid \eta^{e}_t, a_t)$ are \textit{specific} to each environment. Our goal, then, is to learn a \textbf{generalizable policy} $\pi$---that is, one that depends only on $s$.}

%Causal features of action: $s_1, s_2, \dots s_k$
%Noise features: $\eta_1, \eta_2, \dots, \eta_m$

%Does not need to be invariant: $p(\eta_{t+1} \mid \eta_t, a_t)$.
%To discuss: $p(s_{t+1} \mid s_t, a_t)$. It needs to be invariant if we also want to predict $s_{t+1}$ from $s_t$, $a_t$. This could be helpful for state occupancy matching. 

%The emission function $q_e$ and the distribution of $\eta_e$ change between environments while the distribution of $s$ does not. Thus, as illustrated in Figure \ref{fig:block_mdp_imitation}

%Invariant across environment: $p(a\mid s_1, \dots s_k)$

%The environments have a shared dynamics structure and a shared causal structure. 

%Block MDPs represent a special case of our set-up. 

%Inequalities and fairness - try to see if we can do this about

%Block structure assumption: Each observation $x$ uniquely determines its generating state $s$. That is, the observation space $\mathcal{X}$ can be partitioned into disjoint blocks $\mathcal{X}_s$, each containing the support of the conditional distribution $q(\cdot|s)$.

%In the Block MDP setting, the observations might change, but the latent states, dynamics and reward functions are the same. 

\textbf{Structure of Environments:} Second, to learn a policy that depends only on $s$, we must assume that the available training environments are actually different, so that we can learn the invariant state representation using the data from these environments and separate it from the noise representation:

\begin{assumption} \label{assumption:env_interventions}
\squeeze{(Environment Interventions)
Each available training environment $e$ corresponds to a hard \cite{pearl2009causality} or soft \cite{eberhardt2007interventions} intervention on one or more
%variables $x_i$ in the
dimensions of that environment's
observation space (where these
%variables are not
dimensions do not constitute any
causal parents of the demonstrator's actions).}
\end{assumption}

To ensure that a generalizable policy actually exists, Assumption \ref{assumption:latent} requires that $\mathcal{S}$ be non-empty across all environments. Here, to ensure that the space $\mathcal{S}$ can actually be learned, Assumption \ref{assumption:env_interventions} requires that $\mathcal{Z}$ be non-empty across the training environments. Note that we require that the interventions inducing the different environments \textit{not} be on the causal parents of the action, such that Assumption \ref{assumption:latent} is not violated. 

Overall, in our setting $p(x)$ and $p(a\mid x)$ can differ between the multiple environments. However, Assumption
\ref{assumption:latent} enforces that the environments and tasks are the same modulo noise, i.e. that there exists some non-empty $\mathcal{S}$ such that
$p(s)$ and $p(a\mid s)$ are the same between them.  In other words, we have a set of environments that are different (i.e. the dynamics of $x_t$ are different), but the task being performed by the agent is the same (i.e. the dynamics of $s_t$ are the same). This setting applies to the case when lightning conditions in a room are changing, but physical dynamics of the environment are staying the same \cite{zhang2020invariant} or \begin{wrapfigure}{t}{0.55\textwidth}
    \centering
    \vspace{-4mm}
    \includegraphics[width=0.53\textwidth]{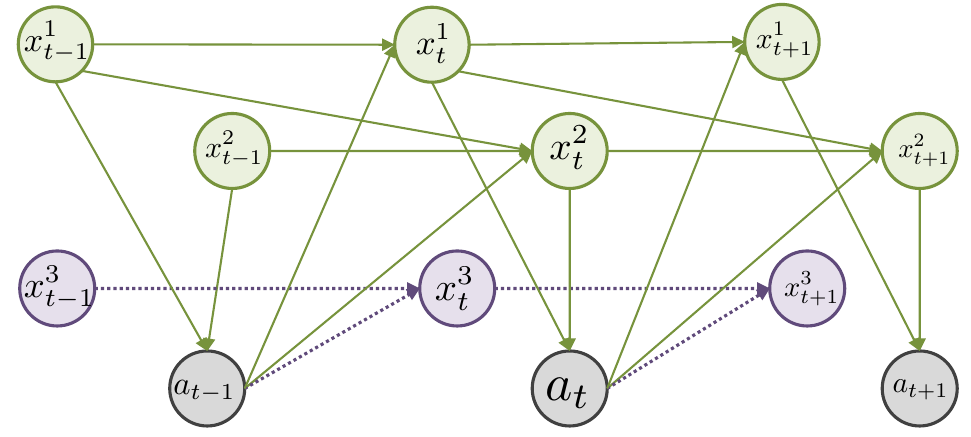}%
    \caption{Causal diagram illustrating temporal dependencies between causal parents of action $\{x_t^1, x_t^2\}$ and noise variables $ \{x^3_t\}$. Different environments are induced by different interventions on the noise variables. }%
    \label{fig:example_causal}
    \vspace{-6mm}
\end{wrapfigure} when weather conditions are changing, but driving behaviour and dynamics are staying the same. We provide additional explanations and definitions of environment interventions in Appendix \ref{apx:scm_interventions}.

Figure \ref{fig:example_causal} shows a simple example where each observation $x_t$ represents a union of the causal parents of the action (state variables) $s_t = \{x_t^1, x_t^2\}$ and the spurious correlations (noise variables) $\eta_t = \{x^3_t\}$. To satisfy Assumption \ref{assumption:env_interventions}, the different environments need to correspond to interventions on $x^{3}_t$. And to satisfy Assumption \ref{assumption:latent}, $x^{1}_t$ and $x^{2}_t$ must not be intervened on. Our aim is to find a representation $s$ of the causal parents $\{x_t^1, x_t^2\}$ of the actions, as well as the mapping between them and the actions $a_t$, that mimics the expert's policy.

Finally, similarly to \cite{zhang2020invariant}, we also assume that the observations $x_t$ at timestep $t$ can only affect the actions $a_t$ and the observations at the next timestep $t+1$:
\begin{assumption} \label{assumption:temp_causal}
(Temporal Causal Mechanism)
Let $x^{i}$ and $x^{j}$ be any two components of the observation $x$ at timestep $t$. Then:
\begin{equation}
    x^{i}_{t+1}  \perp\!\!\!\perp x^{j}_{t+1} \mid x_t, a_t
\end{equation}
\end{assumption}
\squeeze{Note that Assumption \ref{assumption:temp_causal} simply serves to place us within the standard MDP setting: It ensures Markovianity of the temporal transitions, that only the observations $x_t$ at time $t$ will contain the causal parents of the action $a_t$, and that $x_t$ and $a_t$ are the only factors that determine the next observation $x_{t+1}$.}

\section{Invariant Causal Imitation Learning for Domain Generalization}

\squeeze{The goal of our Invariant Causal Imitation Learning (ICIL) algorithm is to learn a representation of the state variables $s$ that is invariant across domains, and an imitation policy $\pi$ that depends on this causal representation and matches the demonstrator's behaviour. We operate in the strictly batch setting, and our aim is for $\pi$ to generalize to unseen environments $e\in\mathcal{E}$ given the above structural assumptions.}

\subsection{Learning Invariant Causal Representations} \label{sec:learning_invariant_causal_rep}

To achieve our goal, we decompose the observations $x^e_t$ in each environment $e$ into a representation $s_t = \phi(x^e_t; \theta_s)$ for the causal features of the action $a_t$, and another representation $\eta^e_t = \mu^{e}(x^e_t; \theta^e_{\eta})$ for the noise variables, where $\theta_s$ and $\theta^e_{\eta}$ are the learnable parameters of $\phi$ and $\eta$. Since the causal parents of the action are invariant across the environments, the state representation model $\phi: \mathcal{X}\rightarrow \mathcal{S}$ is the same across all environments. On the other hand, $\mu^{e}: \mathcal{X} \rightarrow \mathcal{Z}^e$ is environment-specific in order to allow for dynamics mismatch of the noise variables between the different environments.

In order to satisfy the causal diagram in Figure \ref{fig:causal_diagram} and to learn a minimal causal representation, we need the following conditions to be satisfied: (1) $s_t$ should be \textit{invariant} across the environments, (2) $s_t$ and $\eta^{e}_t$ should be \textit{dynamics-preserving}, and (3) $s_t$ and $\eta^{e}_t$ should be \textit{independent} from each other.

\squeeze{Firstly, to fulfill condition (1) we train an environment classifier on the shared state representation $c_{s}: \mathcal{S} \rightarrow |\mathcal{E}_{train}|$, parameterized by $\theta_c$ using the cross-entropy loss. Similarly to \cite{zhang2020invariant}, in order to build a state representation that is invariant across domains, we use an adversarial loss \cite{tzeng2017adversarial} that maximizes the entropy of the classifier: $H(c_s(\phi(x_t; \theta_s); \theta_c))$. This gives us the following practical loss function:}
 \begin{equation}
        \mathcal{L}_{inv}(\theta_s) = \sum_{e\in \mathcal{E}_{train}} \mathbb{E}_{x^{e}_{t} \sim \rho_D^e} - H(c_s(\phi(x^e_t; \theta_s); \theta_c))    
\end{equation}

\squeeze{Out of all possible representations that are invariant, we specifically seek one that also preserves the transition dynamics, fulfilling condition (2). To ensure that the state and noise representations are dynamics-preserving, we also learn the transition dynamics for the state variables $g_s: \mathcal{S} \times \mathcal{A}: \rightarrow \mathcal{S}$, such that $\hat{s}_{t+1} = g_s(s_t, a_t; \theta_{g_s})$ and the environment specific transition dynamics for the noise variables: $g^e_{\eta}: \mathcal{Z}^{e} \times \mathcal{A} \rightarrow \mathcal{Z}^{e}$, such that $\hat{\eta}^{e}_{t+1} = g^{e}_{\eta}(\eta^e_{t}, a_t; \theta^{e}_{g_{\eta}})$. To reconstruct $x_{t+1}$ we also learn $\psi:\mathcal{S} \times \mathcal{Z}^e \rightarrow \mathcal{X}$ such that $\hat{x}^e_{t+1} = \psi(s_{t+1}, \eta^e_{t+1}; \theta_{\psi})$. This yields the following practical loss function:}
\begin{equation}
        \mathcal{L}_{dyn}(\theta_s, \theta_{g_s}, \{\theta^e_{\eta}, \theta^e_{g_{\eta}}\}_{e\in \mathcal{E}_{train}}, \theta_{\psi}) = \sum_{e\in \mathcal{E}_{train}} \mathbb{E}_{x^e_{t+1}, a_{t}, x^e_{t+1} \sim \rho_D^e} \| x^e_{t+1} - \hat{x}^e_{t+1} \|^2
        %\| x_{t+1} - \psi(g_s(\phi(x_t), a_t), g_{\eta}(\mu^{e}(x_t), a_t)) \|^2
\end{equation}
Note that while an alternative approach could consider directly building an invertible mapping  from $x^e$ to $(s, \eta^e)$, the motivation for decoding $s_{t+1}$ and $\eta^e_{t+1}$ into $
\hat{x}^e_{t+1}$ is twofold. In addition to learning dynamics-preserving representations, as we will see in Section 
\ref{sec:matchin_expert_behaviour}, this also allows us to compute the energy of the next state obtained by following the imitation policy and enforcing this to be similar to the distribution of states visited by the expert’s policy.

\begin{figure}[t]
    \centering
	\includegraphics[width=\textwidth]{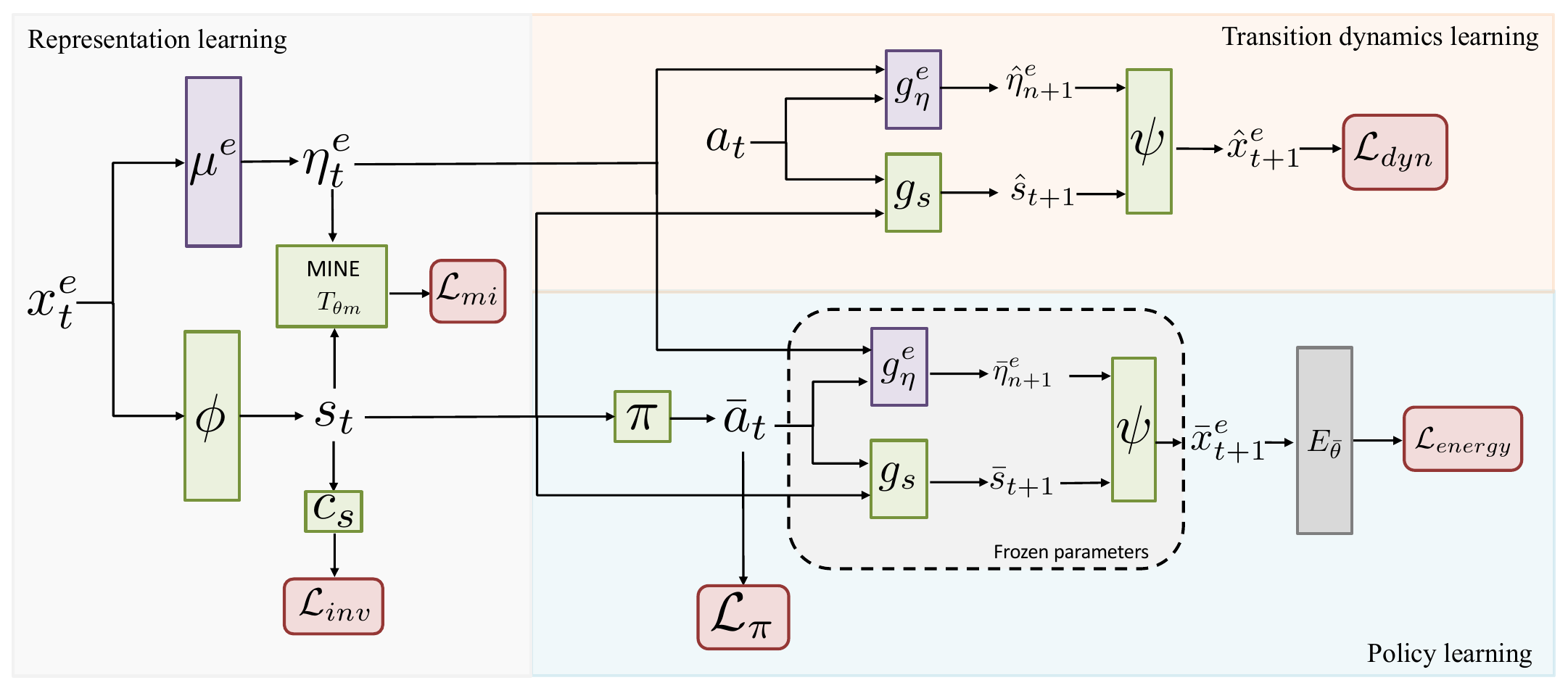}
	\caption{Block diagram of our model. ICIL decomposes the observations $x^e_t$ into an invariant causal representation $s_t$ and an environment specific noise representation $\eta^e$. To obtain an invariant representation, we maximize entropy of an environment classifier that receives as input $s_t$ ($\mathcal{L}_{inv}$). Moreover, the state and noise representations are learnt to be dynamics preserving by minimizing the prediction error of the next observation ($\mathcal{L}_{dyn}$) and independent by minimizing their mutual information ($\mathcal{L}_{mi}$). We learn a generalizable imitation policy that is conditioned on the invariant causal representation ($\mathcal{L}_{\pi}$) and to ensure that the learnt policy matches the distribution of the expert's observations, we minimize the imitator's policy next state energy ($\mathcal{L}_{energy}$).}
	\label{fig:model_diagram}
	\vspace{-2mm}
\end{figure}

Finally, to ensure that the state representation and the noise representation are marginally independent per condition (3), we minimize the mutual information between them. We use the Mutual Information Neural Estimation (MINE) framework \cite{belghazi2018mutual},
which provides a way for estimating the mutual information using neural networks. In particular, MINE uses a neural information measure $I(U, V)$ to approximate the mutual information between random variables $U$ and $V$. Let $T_{\theta_{m}}$ be a statistics network parametrized by $\theta_{m}$. MINE estimates $I(U, V)$ by ascending the gradient of the following:
\begin{equation}
I(U, V) = \sup_{\theta_m} \mathbb{E}_{\mathbb{P}^{(n)}_{UV}} [T_{\theta_m}] - \log (\mathbb{E}_{\mathbb{P}^{(n)}_U \otimes \mathbb{P}^{(n)}_{V}} [e^{T_{\theta_m}} ]) = \sup_{\theta_m} I(U, V; \theta_m)
\end{equation}
where $\mathbb{P}_{UV}$ is the joint measure of $(U, V)$ and  $\mathbb{P}_{U} = \int_{\mathcal{V}} dP_{UV}$, $\mathbb{P}_{V} = \int_{\mathcal{U}} dP_{UV}$ are the marginal distributions. $\mathbb{P}^{(n)}$ denotes the empirical distribution associated with $n$ i.i.d samples. As noted in \cite{belghazi2018mutual}, the neural information measure $I(U, V)$ can approximate the mutual information with arbitrary accuracy. We therefore add the following practical loss function to our optimization objective, which seeks to minimize the mutual information between the state representation and noise representation:
\begin{equation}
        \mathcal{L}_{mi}(\theta_s, \{\theta^e_{\eta}\}_{e\in \mathcal{E}_{train}}) = \sum_{e\in \mathcal{E}_{train}} \mathbb{E}_{x^e_{t} \sim \rho_D^e} I(\phi(x^{e}_t; \theta_s), \mu(x^{e}_t; \theta^e_{\eta}); \theta_m)   
\end{equation}
Note that the parameters $\theta_m$ of the statistics network $T_{\theta_m}$ used for computing the mutual information are updated through gradient ascent on $I(U,V;\theta_{m})$.
%{\color{red}(should this be $I(U,V;\theta_{m})$?)}.

%Moreover, to ensure that the state representation is invariant across domains and that the state and noise variables are disentangled we add two additional components to our model. 

%What we really want is s and eta such that some conditions are fullfiled. If the condition is fulfilled we need specific loss functions. 

%There are lots of spaces we can decompose the state that are invariant; how to chose? we pick one that respects the dynamics.

%Out of the representaitons that are invariant we want the ones that respect the dynamics. 
%(a) invariant
%(b) preserve the dynamics
%(c) if it's noise - no mutual information (ensure disentanglement)

\subsection{Matching Expert Behaviour in a Strictly Batch Setting} 
\label{sec:matchin_expert_behaviour}

On the basis of the causal representation $s$, we shall learn a generalizable policy $\pi$ (parameterized by $\theta_{\pi}$) in the strictly batch setting, such that it matches the demonstrator's behaviour. To begin, we first condition $\pi$ on the representation $s_t$ and minimize the negative log-likelihood of the expert's actions:
    \begin{equation}
        \mathcal{L}_{\pi}(\theta_{\pi}, \theta_s) = \sum_{e\in \mathcal{E}_{train}} - \mathbb{E}_{x^{e}_t, a_t \sim \rho_D^e} \log \pi(a_t \mid \phi(x_t^e; \theta_{s}); \theta_{\pi})
    \end{equation}
However, having only this objective corresponds to performing behaviour cloning, which has well-known limitations \cite{melo2010learning, ross2010efficient, ross2011reduction, piot2014boosted}. To mitigate compounding error, we want some form of added regularization to incentivize the imitation policy to stay within the distribution of the expert's observations.

In the online setting, a popular approach is to make sure that the rollout distribution of the imitating policy matches the rollout distribution of the expert's policy---for instance, by minimizing some form of divergence between their induced occupancy measures. However, this requires interactive access to the real environment or simulator to perform rollouts of intermediate policies---which is not possible in our setting. Instead, we propose a method that takes advantage of the learnt transition dynamics. For any current observation $x_{t}\sim \rho_D$, we shall encourage the next observation $\bar{x}_{t+1}$ obtained by following the imitation policy $\bar{a}_t \sim \pi(\cdot \mid x_{t})$ to remain within the occupancy measure of the expert.

Consider approximating the expert's occupancy measure using an Energy Based Model (EBM) such that $\rho_{D}(x) = \frac{\exp(-E_{\bar{\theta}}(x))}{Z(\bar{\theta})}$ where the function $E_{\bar{\theta}}(x): \mathcal{X} \rightarrow \mathbb{R}$ is the energy function and $Z(\bar{\theta}) = \int_{x} -E_{\bar{\theta}}(x) dx$ is the partition function. We parameterize $E_{\bar{\theta}}$ by a neural network. It is not possible to train the EBM directly through maximum likelihood because $Z(\bar{\theta})$ involves integrating over the entire input domain of $x$ which is impractical. Instead, we use contrastive divergence to pre-train the energy function  $E_{\bar{\theta}}$ \cite{hinton2002training, du2019implicit}.  Contrastive divergence lowers the energy of the observations coming from the expert's occupancy distribution and increases the energy of the observations outside of the expert's occupancy distribution. Refer to Appendix \ref{apx:energy_model} for details on how we train the EBM. 

%Lengevin dynamics. \cite{welling2011bayesian}
To incentive the imitation policy to stay within the distribution of the expert's observations, we train it to minimize the energy of the next observation obtained by following $\pi$ given the current observation:

\begin{equation}
    \mathcal{L}_{energy}(\theta_{\pi}; \theta_s, \theta_{g_s}, \{\theta^e_{\eta}, \theta^e_{g_{\eta}}\}_{e\in \mathcal{E}_{train}}, \theta_{\psi}) = \sum_{e\in \mathcal{E}_{train}} \mathbb{E}_{\substack{x^e_{t} \sim \rho_D^e \\ s_t = \phi(x^e_t), \eta^{e}_t = \mu^e(x^e_t) \\ \bar{a}_t \sim \pi(\cdot \mid s_t) \\ \bar{x}_{t+1} = \psi(g_s(s_t, \bar{a}_t), g^e_{\eta}(\eta^{e}_t, \bar{a}_t))  }} E_{\bar{\theta}}(\bar{x}^e_{t+1})
\end{equation} 
This effectively assigns a high ``reward'' to the imitation policy for staying within high-density areas of the expert's occupancy measure, and a low ``reward'' for straying from it. This can be seen as an adaptation of online imitation methods \cite{reddy2019sqil, liu2020energy} where the expectation would be instead over $x_{t} \sim \rho_{\pi}$.

We illustrate in Figure \ref{fig:model_diagram} all of the components of the our ICIL model. Further details and the full algorithm for optimizing ICIL can be found in Appendix \ref{apx:full_algorithm}. 

%In the online setting, we would have . This would be equivalent to methods in the online setting which give a high reward reward for staying withing the expert's distribution. ]

%Say that in online setting we would have rho pi for $x_t$ and then say that would be equivalent to other imitation learning methods in the online setting.

%EDM, RCAL are biased. ValueDICE is discounted and we do not want to discount in healthcare. So we do not want to do distribution matching. We do something that is quite simple (we match energy for one step). 

%SQIL is 1 and 0 rewards. Give high reward for staying whitin expert distribution. Offline version of that - 1 step. In energy based imitation learning 

%Write equation for what we could do if this is online. 

%\begin{algorithm}[t]
%\begin{algorithmic}[1] 
%\item For
%\caption{Invariant Causal Imitation Learning} \label{alg:cirl}
%\end{algorithmic} 
%\end{algorithm}

\section{Experiments}

We perform experiments on OpenAI gym tasks \cite{brockman2016openai} and on an ICU dataset from the MIMIC III database \cite{johnson2016mimic}. In both cases, we generate data from multiple domains by augmenting the feature space with noise variables (spurious correlations).

\textbf{Benchmarks} We compare ICIL\footnote[1]{The code for ICIL can be found at \url{https://github.com/vanderschaarlab/mlforhealthlabpub} and at \url{https://github.com/ioanabica/Invariant-Causal-Imitation-Learning}.} against standard methods for strictly batch imitation learning: Behaviour Cloning (BC) \cite{pomerleau1991efficient};  
Reward-regularized Classification for Apprenticeship Learning (RCAL), which incorporates dynamics-awareness through a sparsity regularization on the implied rewards \cite{piot2014boosted}; ValueDICE (VDICE) \cite{kostrikov2019imitation}, which uses an off-policy objective to estimate distribution ratios needed for distribution matching; as well as Energy-based Distribution Matching (EDM) \cite{jarrett2020strictly}, which jointly learns the imitator policy with an energy model of its state distributions. These methods seek to find a policy that approximately matches the expert's behaviour from a single environment, and were not designed with generalization in mind. Hence we augment these benchmark by using the IRMv1 objective \cite{arjovsky2019invariant} in conjunction with their originally defined imitation risk to obtain the additional benchmarks: BC-IRM, RCAL-IRM, VDICE-IRM, and EDM-IRM. More details about how we used the invariance-based penalty from IRM \cite{arjovsky2019invariant} to augment these existing methods such that they may learn generalizable policies can be found in Appendix \ref{apx:irm_imitation}. Implementation details about all benchmarks and the hyperparameter settings used can be found in Appendix \ref{apx:experiments_details}.

\begin{figure}%
	\subfloat[Acrobot]{\includegraphics[width=3.5cm]{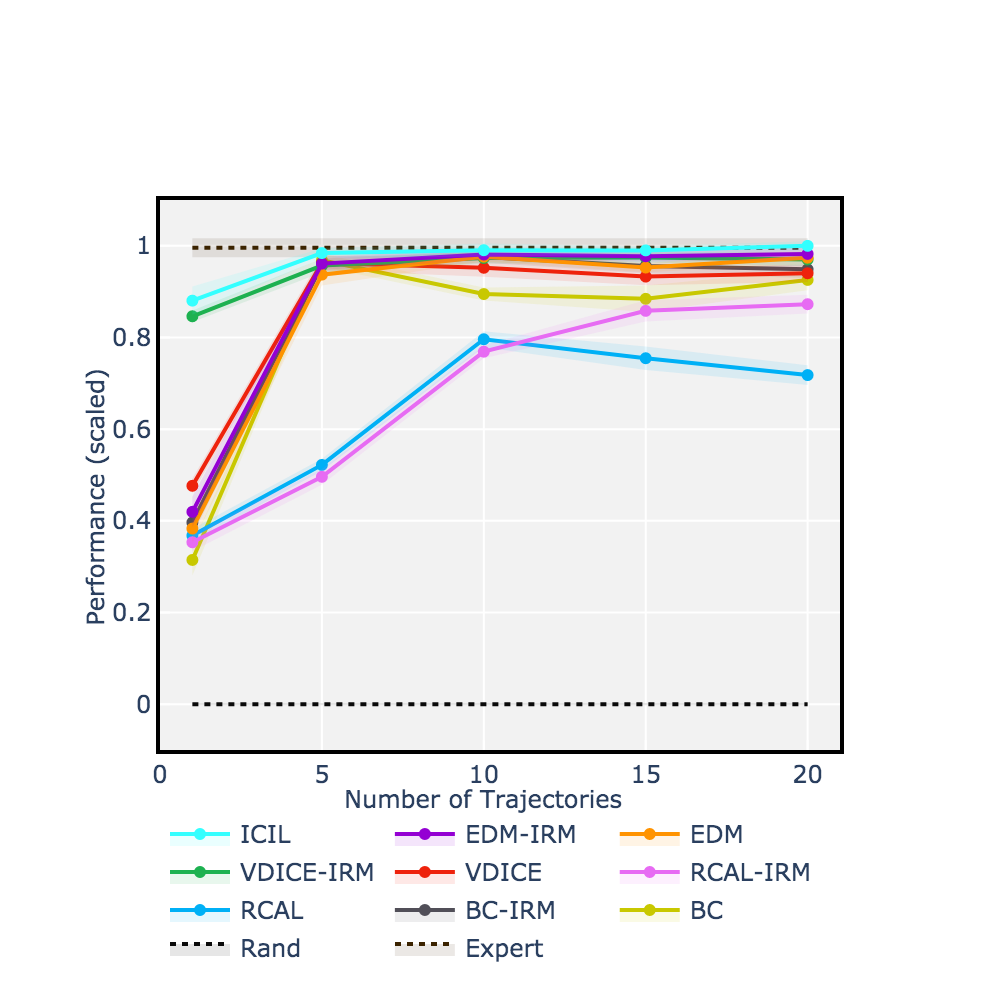}}%
	\subfloat[Cartpole]{\includegraphics[width=3.5cm]{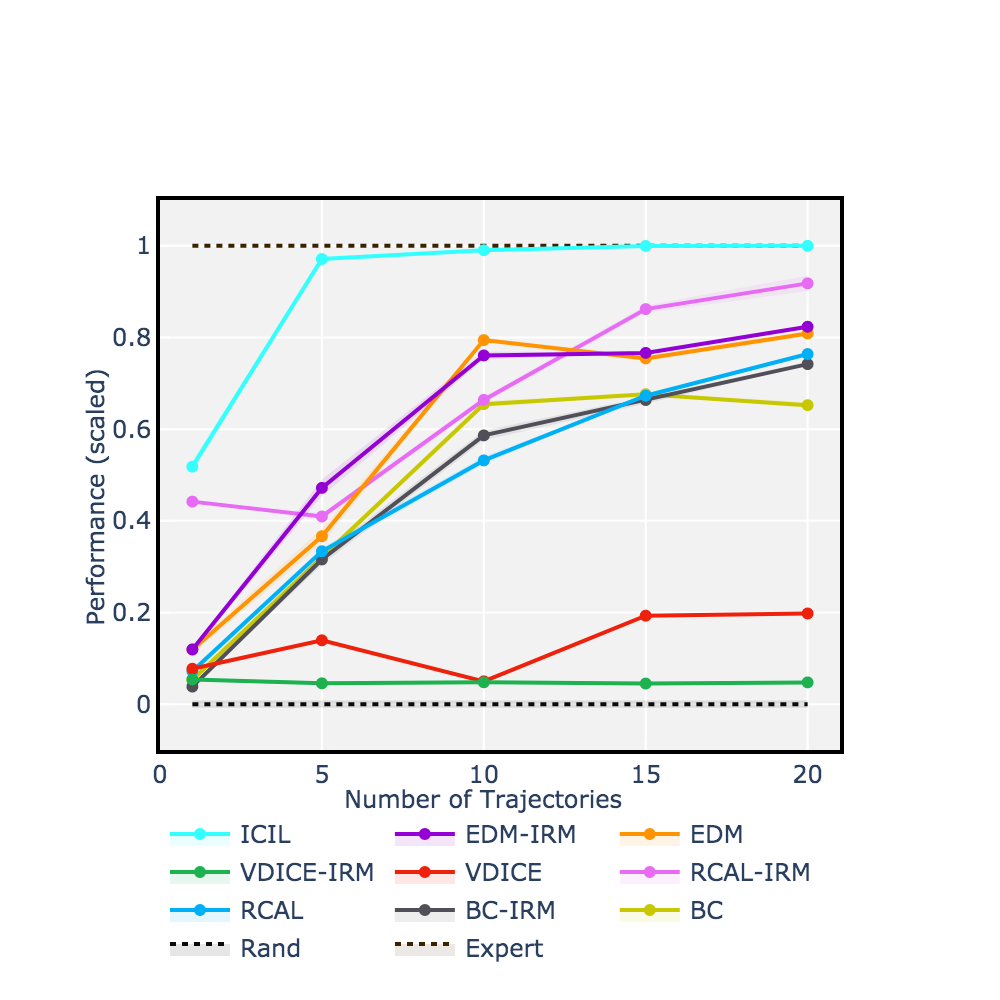}}%
	\subfloat[LunarLander]{\includegraphics[width=3.5cm]{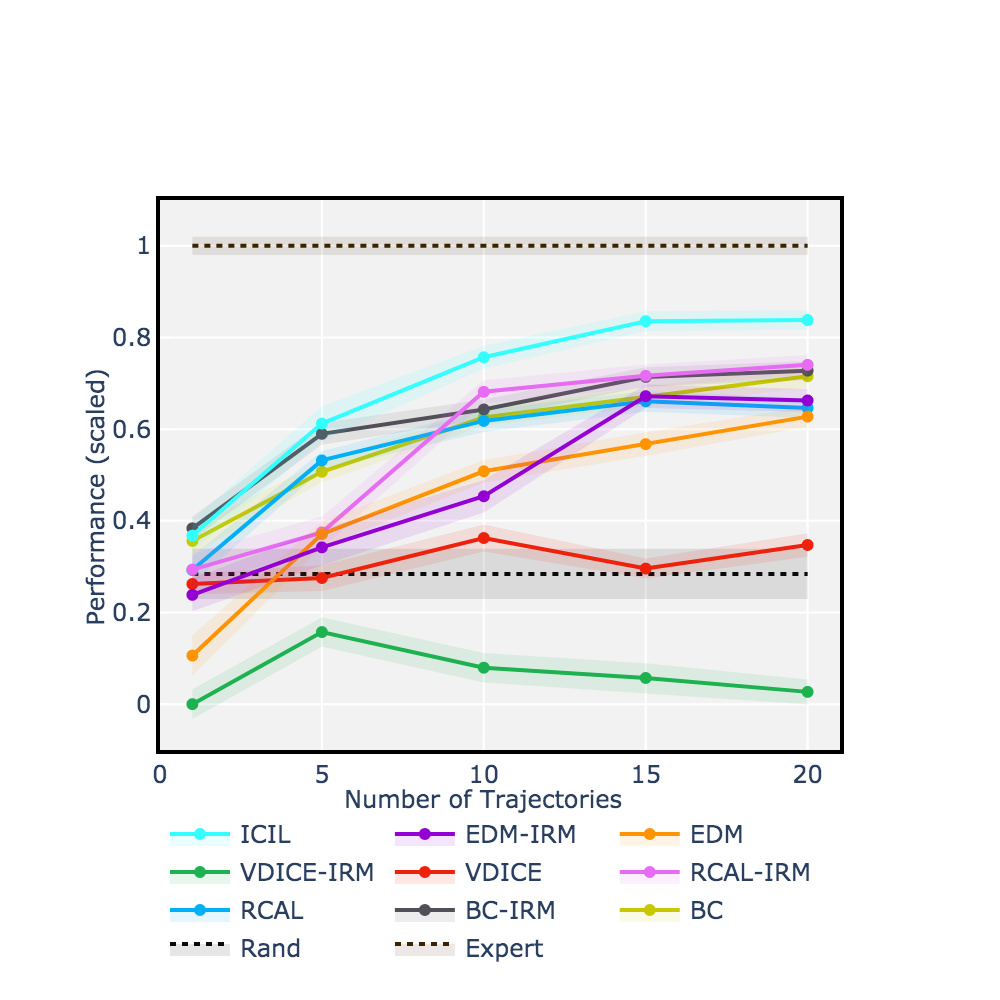}}%
	\subfloat[BeamRider]{\includegraphics[width=3.5cm]{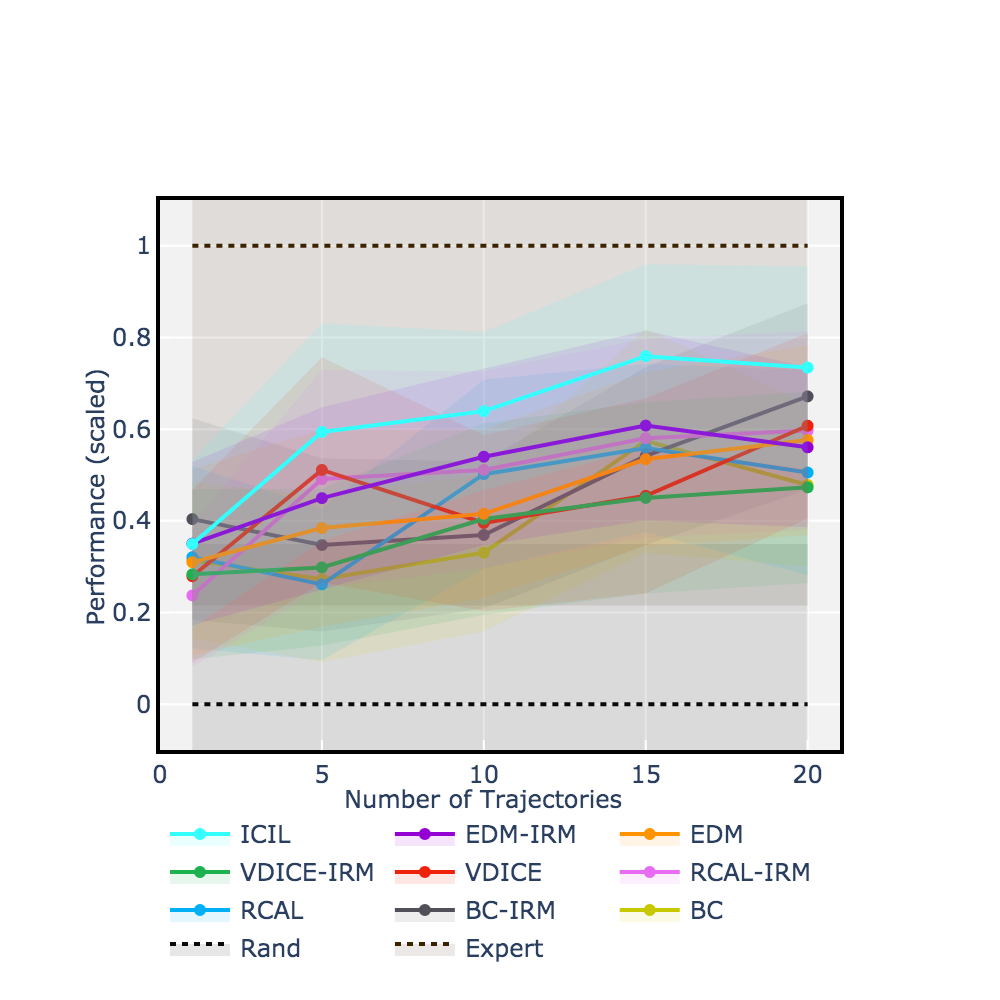}}%
	\caption{Evaluation on OpenAI gym environments. $x$-axis indicates the number of trajectories (in $\{1, 5, 10, 15, 20\}$) with expert demonstrations from each training environment given as input to each benchmark and $y$-axis represents the average return of the learnt imitation policy on the test environments, scaled between 1 (expert performance) and 0 (random policy performance). }    \label{fig:results_gym}%
	\vspace{-4mm}
\end{figure}

\subsection{Evaluation on OpenAI Gym} \label{sec:eval_openai_gym}
We perform experiments on the following control tasks from OpenAI gym \cite{brockman2016openai}: Acrobot \cite{geramifard2015rlpy}, Cartpole \cite{barto1983neuronlike}, LunarLander \cite{brockman2016openai} and BeamRider \cite{bellemare2013arcade}. For each task, we use pre-trained RL agents from RL Baselines Zoo \cite{rl-zoo} and Stable OpenAI Baselines \cite{stable-baselines} to obtain expert policies. We then follow an approach similar to the one in \cite{zhang2020invariant} to obtain datasets with demonstrations from the expert in two different environments. In particular, for Acrobot \cite{geramifard2015rlpy},  Cartpole \cite{barto1983neuronlike} and LunarLander \cite{brockman2016openai} we add spurious correlations to the state space of each control task and an environment identifier. The spurious correlations in each environment are different multiplicative factors of a subset of variables in the original state space. The invariant causal state is represented by the original variables in the state space of each control task. We train the benchmarks on demonstrations from two environments with $1\times$ and $2\times$ multiplicative factors for the spurious correlations and we test on an environments with multiplicative factors sampled from $\mathcal{U}(-1, 1)$.  For BeamRider \cite{bellemare2013arcade}, similarly to \cite{zhang2020invariant}, different camera angles are used for the training and testing environments. In particular, we use two training environments where the game frames are rotated by 10 degrees to the left and to the right respectively, while the test environment has no rotation. The rotation is applied to the entire frame for all trajectories in each environment. However, note that, despite the rotation, the dynamics for the state variables  and how they influence the action stay the same. Further details about the train and test environments can be found in Appendix \ref{apx:experiments_details}.

We vary the number of demonstrated trajectories from each environment that we give as input to each benchmark and we evaluate them on the average return obtained by deploying the learnt imitation policies on the test environment.
Figure \ref{fig:results_gym} shows the mean results and standard errors obtained across 10 runs where for each run we train the benchmarks on different trajectories from the train environments and we evaluate on a test environment with newly sampled multiplicative factors for computing the spurious correlations. We notice that our method consistently outperforms the benchmarks and is capable of generalizing better to the unseen target environments. Moreover, we generally found that using the IRMv1 objective \cite{arjovsky2019invariant} together with existing methods for strictly batch imitation learning did not improve performance and resulted in more unstable training. For additional results, see Appendix \ref{apx:additional_experiments} where we perform ablation studies to investigate the impact of the different terms in the loss function used to train ICIL on overall performance, compare performance on train vs. test environments and also evaluate robustness to increasing the size of the spurious correlations.

%Add spuriously correlated dimensions which are a multiplicative factor of the original state space and add an environment identifier + add noise on the true causal state. Change multiplicative factor across environments. -- this means that the noise variables change at every timestep 

\subsection{Evaluation on MIMIC-III} 

We also perform experiments on a healthcare dataset with Intensive Care Unit (ICU) patients extracted from the Medical Information Mart for Intensive Care (MIMIC-III) database \cite{johnson2016mimic}. 
The dataset consists of trajectories of clinical measurements (e.g. heart rate, respiratory rate) recorded every hour. The aim is to learn a generalizable policy for the action of putting patients on the mechanical ventilator.

We define three environments, two for training and one for testing, each consisting of 2000 independent patient trajectories from MIMIC-III. We augment the original feature space by adding spurious correlations (noise variables) that are the same as the expert actions with probabilities $p=0.1$ and $p=0.2$ in the training environments and with probability $p=0.8$ in the testing environment. 

\begin{table}
	\centering
	\begin{tabular}{lccc}
			\toprule
			 & \multicolumn{3}{c}{\textbf{Mechanical ventilator}}  \\
			Benchmark & \multicolumn{1}{c}{ACC} & \multicolumn{1}{c}{AUC} & \multicolumn{1}{c}{APR} \\
			\midrule
			BC & $0.783\pm0.001$ & $0.762\pm0.002$ & $0.692\pm0.001$  \\
			RCAL & $0.790\pm0.002$ & $0.771\pm0.002$ & $0.697\pm0.002$  \\
			VDICE & $0.794\pm0.001$ & $0.784\pm0.001$ & $0.716\pm0.001$ \\
			EDM & $0.786\pm0.003$ & $0.741\pm0.011$ & $0.682\pm0.005$ \\
			\midrule
			BC-IRM & $0.791\pm0.002$ & $0.767\pm0.003$ & $0.696\pm0.002$ \\
			RCAL-IRM & $0.789\pm0.002$ & $0.766\pm0.003$ & $0.694\pm0.003$ \\
			VDICE-IRM & $0.766\pm0.001$ & $0.730\pm0.001$ & $0.694\pm0.001$  \\
			EDM-IRM & $0.781\pm0.004$ & $0.717\pm0.015$ & $0.673\pm0.007$  \\  
			\midrule 
			ICIL & $\bm{0.855\pm0.003}$ & $\bm{0.856\pm0.004}$ & $\bm{0.789\pm0.004}$ \\
			\bottomrule
	\end{tabular}
	\vspace{0.1cm}
	\caption{Evaluation on MIMIC-III in terms of action-matching. We compare the actions selected by the benchmark imitation policies with the ones from the clinical expert policy in the test environment and report the accuracy (ACC), the area under the receiving operator characteristic curve (AUC) and the area under the the precision-recall curve (APR).}
	\label{tab:mimic}
	\vspace{-0.3cm}
\end{table}
In a real setting, such spurious correlations are commonplace. For instance, consider some hospitals (i.e. training environments) where selection bias is present, such that patients with a certain otherwise irrelevant comorbidity happen to receive a treatment more often \cite{taylor2010blood, desautels2017prediction, soo2019describing, subbaswamy2021evaluating}. However, learning an imitation policy that takes into account such a comorbidity when assigning the patient's treatment would fail to generalize to hospitals where fewer patients suffer from this comorbidity but should still receive the treatments. More details about the dataset can be found in Appendix \ref{apx:experiments_details}.  

Since MIMIC-III is an entirely offline dataset, it is not possible to compute average returns for running the policies in the test environment. Instead, we evaluate the benchmarks in terms of action matching on the test environment. We report in table \ref{tab:mimic}
the mean accuracy (ACC), the mean area under the receiving operator characteristic curve (AUC), the mean area under the the precision-recall curve (APR) and their standard deviations over 10 runs. We notice that ICIL learns a policy that best discards the spurious correlations present in the training environment to learn a generalizable policy for putting patients on the mechanical ventilator that best matches the expert's actions on the test environment.

\section{Discussion} \label{sec:discussion}

In this paper, we tackle the problem of learning generalizable imitation policies in the strictly batch setting. Our ICIL model leverages ideas from causality and learns an invariant state representation that minimizes the presence of spurious correlations. By conditioning the imitation policy on this state representation, we obtain a policy that generalizes to environments with the same shared latent structure, but with different noise distribution and dynamics. ICIL also matches expert behaviour by incentivizing the learnt imitation policy to stay within the expert's observations distribution. 

In terms of limitations, we believe that future work should consider providing theoretical insights and error bounds on the generalization error. In addition, to be able to learn an invariant state representation, our method requires demonstrated trajectories from at least two training environments with different interventions on the noise variables (spurious correlations), and the method cannot be used if such data is not available in practice. Finally, we bear in mind that---as with any other imitation learning method that aims to match the expert's policy---ICIL can have potential negative societal impacts if the expert's policy is flawed in the first place. Thus, in sensitive applications such as clinical decision support, care must be taken to prevent potentially negative feedback loops.
% This can create a negative feedback loop in clinical decision support systems where if we mimic slightly incorrect clinical expert decisions, when deploying such a behaviour policy in a new environment it will perpetuate the flaws. 

%in terms of potential negative societal impacts, ICIL aims to directly match the expert's policy. If the expert's policy is biased/unfair and uses sensitive attributes as one of the causes to the action, the policy learnt by our model will perpetuate such biases. Because of this, it would be important to first audit the expert's policies to ensure that it is free from such biases before using our method to match it and deploy it in new environments. 

\section*{Acknowledgments}
We would like to thank the reviewers for their valuable feedback. The research presented in this paper was supported by The Alan Turing Institute, under the EPSRC grant EP/N510129/1, by Alzheimer’s Research UK (ARUK), by the US Office of Naval Research (ONR), and by the National Science Foundation (NSF) under grant number 1722516.

\bibliography{refs.bib}
\bibliographystyle{unsrt}

\newpage
\appendix

\section{Structure of environment and interventions} \label{apx:scm_interventions}

Similarly to \cite{arjovsky2019invariant}, we consider that all environments have the same underlying Structural Causal Model (SCM) and that the different environments correspond to different interventions on the SCM. We provide here the formal definition for SCMs and interventions. 

\begin{definition} (Structural Causal Model)  \cite{arjovsky2019invariant}: A structural causal model (SCM) $\mathcal{C} = (S, N)$ governing the random vector $X = (X_1, \dots X_m)$ is a collection $S$ of $m$ assignments:
\begin{equation}
    S_j: X_j \leftarrow f_j(Pa(X_j), N_j), \text{ for } j = 1, \dots m
\end{equation}
where $Pa(X_j) \subseteq \{X_1, X_2, \dots X_m\} / \{X_j\}$ are the parents of $X_j$ and the $N_j$ are the independent noise variables. We say that $X_i$ causes $X_j$ if $X_i \in Pa(X_j)$. 
\end{definition}

\begin{definition} (Intervention) \cite{arjovsky2019invariant}: Consider a SCM $\mathcal{C} = (S, N)$. An intervention $e$ on $\mathcal{C}$ consists of replacing one or several of its structural equations to obtain an intervened SCM $\mathcal{C}^{e} = (S^e, N^e)$ with structural equations:  
\begin{equation}
    S^e_j: X^e_j \leftarrow f_j(Pa(X^e_j), N^e_j), \text{ for } j = 1, \dots m
\end{equation}
The variable $X^{e}$ is intervened on if $S_i \neq S_i^e$ or $N_i \neq N_i^e$.
\end{definition}

In our setting, the variables forming the SCM are the different observations and actions at each timestep. Moreover, Assumption \ref{assumption:temp_causal} that ensures the Markovianity of the temporal transitions restrict the relationships that can be present in the SCM. In addition, Assumption \ref{assumption:env_interventions} requires that the interventions to not be on the causal parents of the action.

Soft interventions \cite{eberhardt2007interventions} do not remove any edges in the causal graph induced by the SCM, but instead modify the conditional probability distributions of the variables intervened on. On the other hand, hard interventions \cite{pearl2009causality} on a variable remove all incoming edges from the parents of that variables.

\newpage

\section{Train energy based model} \label{apx:energy_model}

We train the energy-based model for the expert demonstrations using Persistent Contrastive Divergence \cite{du2019implicit}. To sample from the energy-based model, we use Markov Chain Monte Carlo using Langevin Dynamics\cite{welling2011bayesian}. Algorithm \ref{alg:ebm} outlines the method used to learn the energy-based model $E_{\bar{\theta}}$ of the expert observations. See \cite{du2019implicit} for details on training EBMs in this manner.

\begin{algorithm}[h]
\begin{algorithmic}[1] 
\State \textbf{Input:} Dataset with expert demonstrations $\mathcal{D}$, number of steps $K$, step size $\alpha$, noise variance $\sigma$, Mini-batch size $N$
\State \textbf{Initialize:} energy-based model parameters $\bar{\theta}$, buffer $\mathcal{B} \leftarrow \varnothing$
\While{not converged}
    \State Sample $N$ positive samples from expert demonstrations $x_i^{+} \sim \mathcal{D}$
    \State Sample initial negative samples: $x_{i}^{0} \sim \mathcal{B}$ with $95\%$ probability and $x_{i}^{0} \sim \mathcal{U}(-1, 1)$ otherwise
    \For{sample step $k=1$ to $K$} \Comment{Generate sample via Langevin dynamics}
    \State $\tilde{x}^{k}_i \leftarrow \tilde{x}_i^{k-1} - \alpha \cdot \nabla_{x} E_{\bar{\theta}}(\tilde{x}_i^{k-1}) + \omega, \text{where } \omega \sim \mathcal{N}(0, \sigma)$, $\forall i\in \{1, \dots N\} $
    \EndFor
    \State $x_{i}^{-} = \Omega(x_i^{K})$, $\forall i\in \{1, \dots N\} $ \Comment{$\Omega$: stop gradient operator}
    \State Contrastive divergence loss $\mathcal{L}_{CD} = \frac{1}{N} \sum_{i} E_{\bar{\theta}}(x_{i}^{+}) - E_{\bar{\theta}}(x_{i}^{-})$
    \State Regularization loss: $\mathcal{L}_{RG} = \frac{1}{N} \sum_{i} E_{\bar{\theta}}(x_{i}^{+})^2 + E_{\bar{\theta}}(x_{i}^{-})^2$
    \State Update parameters $\bar{\theta}$ by backpropagating $\nabla_{\bar{\theta}} (\mathcal{L}_{CD} + \mathcal{L}_{RG})$
    \State Add samples to buffer: $\mathcal{B}\leftarrow \mathcal{B} \cup \{x_{i}^{-}\}_{i=1}^{N}$
\EndWhile
\caption{Learning energy-based model $E_{\bar{\theta}}$ of expert demonstrations} \label{alg:ebm}
\end{algorithmic} 
\end{algorithm}

For the Acrobot \cite{geramifard2015rlpy}, CartPole 
\cite{barto1983neuronlike}, LunarLander \cite{brockman2016openai} control tasks and the experiments on the healthcare dataset extracted from MIMIC III 
\cite{johnson2016mimic}, we use a neural network with 2 fully-connected hidden layers of size $64$ and with ReLU activation to define $E_{\bar{\theta}}$. For Acrobot \cite{geramifard2015rlpy}, CartPole 
\cite{barto1983neuronlike}, LunarLander \cite{brockman2016openai}, we set the hyperparameters to number of steps $K=100$, step size $\alpha = 0.01$, noise variance $\sigma = 0.01$ and mini-batch size $N=64$. We optimize $\bar{\theta}$ by using the Adam optimizer for 1000 training iterations with the learning rate set to $0.001$. For the experiments on the healthcare dataset extracted from MIMIC III \cite{johnson2016mimic}, we use the following hyperparameters for the number of steps $K=50$, step size $\alpha = 0.01$, noise variance $\sigma = 0.01$ and mini-batch size $N=128$. We optimize $\bar{\theta}$ by using the Adam optimizer for 1000 training iterations with the learning rate set to $0.0005$.

To define $E_{\bar{\theta}}$ for the BeamRider Atari environment \cite{bellemare2013arcade} we use a convolutional neural network with 3 convolutional layers with 32-64-64 filters, followed by a fully connected layer of size 64, with all layers followed by ReLU activations. The hyperparameters are set as follows:  number of steps $K=100$, step size $\alpha = 0.01$, noise variance $\sigma = 0.01$ and mini-batch size $N=64$. $\bar{\theta}$ is optimized by using the Adam optimizer for 1000 training iterations with the learning rate set to $0.001$.

\newpage
\section{Algorithm} \label{apx:full_algorithm}

Algorithm \ref{alg:icil} provides the pseudo-code for training Invariant Causal Imitation Learning (ICIL). 
\begin{algorithm}[h]
\begin{algorithmic}[1] 
\State Input: Dataset with expert demonstrations $\mathcal{D}$, learning rate $\lambda$, mini-batch size $N$
\State Initialize: $\theta_s, \theta_{g_s}, \{\theta^e_{\eta}, \theta^e_{g_{\eta}}\}_{e\in \mathcal{E}_{train}}, \theta_{\psi}, \theta_c, \theta_{\pi}, \theta_m$ 
\While{not converged}
    \State Sample mini-batch of N demonstrations $(x^{e_i}_{i}, a_i, x^{e_i}_{i+1}) \sim \mathcal{D}$
    \State Sample permutation of $[N] = \{1, \dots N\}$ from uniform distribution over the set of all permutations $S_N$: $\kappa \sim \mathcal{U}(S_N)$ 
    \State \textbf{ICIL update}: 
    \State Invariance loss: $\mathcal{L}_{inv} = \frac{1}{N} \sum_{i=1}^{N} - H(c_s(\phi(x_i^{e_i})))$ 
    \State Transition dynamics loss: 
    \begin{equation*}
        \mathcal{L}_{dyn} = \frac{1}{N} \sum_{i=1}^{N} \|x_{i+1}^{e_i} - \psi(g_s(\phi(x_i^{e_i}), a_i), g_{\eta}^{e_i}(\mu^{e_i}(x_i^{e_i}), a_i)) \|^2
    \end{equation*}
    \State Mutual information loss:
    \begin{equation*}
        \mathcal{L}_{mi} = \sum_{i=1}^{N} T_{\theta_m} (\phi(x^{e_i}_i), \mu(x^{e_i}_i)) - \log (\sum_{i=1}^{N} \exp T_{\theta_m} (\phi(x^{e_i}_i), \mu(x^{e_{\kappa(i)}}_{\kappa(i)})))
    \end{equation*}
    \State Policy loss:
    \begin{equation*}
        \mathcal{L}_{\pi} = \frac{1}{N} \sum_{i=1}^{N} \text{Cross entropy}(\pi(\cdot \mid \phi(x_i^{e_i})), a_i)
    \end{equation*}
    \For{$i=1\dots N$}
    \State $\bar{a}_i \sim \text{Gumbel Softmax} (\pi(\cdot \mid \phi(x_i^{e_i}))) $
    \EndFor
    \State Next state energy loss:
    \begin{equation*}
        \mathcal{L}_{energy} = \frac{1}{N} \sum_{i=1}^{N} E_{\bar{\theta}}( \psi(g_s(\phi(x_i^{e_i}), \bar{a}_i), g_{\eta}^{e_i}(\mu^{e_i}(x_i^{e_i}), \bar{a}_i)))
    \end{equation*}
    \Statex
    \State Parameters update: 
    \State $\theta_s \leftarrow \theta_s - \lambda \nabla_{\theta_s} (\mathcal{L}_{inv} + \mathcal{L}_{dyn} + \mathcal{L}_{mi} +  \mathcal{L}_{\pi})$
     \State $\theta_{g_s} \leftarrow  \theta_{g_s} - \lambda \nabla_{\theta_s} \mathcal{L}_{dyn}$
     \State $\theta_{\psi} \leftarrow \theta_{\psi} - \lambda \nabla_{\theta_{\psi}} \mathcal{L}_{dyn}$
     \For{$e \in \mathcal{E}_{train}$}
        \State $\theta^e_{\eta} \leftarrow  \theta^e_{\eta} - \lambda \nabla_{\theta^e_{\eta}} (\mathcal{L}_{dyn} + \mathcal{L}_{mi})$
        \State $\theta^e_{g_{\eta}} \leftarrow  \theta^e_{g_{\eta}} - \lambda \nabla_{\theta^e_{g_{\eta}}} \mathcal{L}_{dyn}$
    \EndFor
    \State $\theta_{\pi} \leftarrow \theta_{\pi} - \lambda \nabla_{\theta_{\pi}} (\mathcal{L}_{\pi} + \mathcal{L}_{energy})$
    \Statex
    \State \textbf{Environment classifier update:} \Comment{Used to define the invariance loss.}
    \State $\mathcal{L}_c = \frac{1}{N}\sum_{i=1}^N \text{Cross entropy}(c_s(\phi(x_i^{e_i}), e_i)$
    \State $\theta_c \leftarrow \theta_c - \lambda \nabla_{\theta_c} \mathcal{L}_c$
    \Statex
    \State \textbf{Mutual information (MINE) update:} \Comment{Used to define the mutual information loss.}
    
    \State Update $T_{\theta_m}$ by ascending the gradient of the mutual information loss:
    $\theta_m \leftarrow \theta_m + \nabla_{\theta_m} \mathcal{L}_{mi}$
\EndWhile
\State \textbf{Output:} Learnt parameters $\theta_s, \theta_{g_s}, \{\theta^e_{\eta}, \theta^e_{g_{\eta}}\}_{e\in \mathcal{E}_{train}}, \theta_{\psi}, \theta_c, \theta_{\pi}, \theta_m$ 
\caption{Invariant Causal Imitation Learning} \label{alg:icil}
\end{algorithmic} 
\end{algorithm}

\newpage
\section{Causal features for imitation using invariant risk minimization} \label{apx:irm_imitation}

In the supervised learning setting, Invariant Risk Minimization (IRM) \cite{arjovsky2019invariant} leverage data from multiple domains to learn a data representation $\Phi: \mathcal{X}\rightarrow \mathcal{H}$ that elicits an invariant predictor $w:\mathcal{H}\rightarrow \mathcal{Y}$ across the different environments, where $\mathcal{X}$ and $\mathcal{Y}$ are the input and output spaces respectively. The training data from each environment $e \in \mathcal{E}$ corresponds to different interventions on the data generating process and $R^{e}$ corresponds to the empirical risk of the classifier in each domain. We use the following formal definition from \cite{arjovsky2019invariant} to describe the characteristics we want from the learnt representation. 

\begin{definition} \cite{arjovsky2019invariant}: We say that a data representation $\Phi: \mathcal{X}\rightarrow \mathcal{H}$ elicits an invariant predictor across environments $\mathcal{E}$ if there is a classifier $w:\mathcal{H}\rightarrow \mathcal{A}$ simultaneously optimal for all environments, that is, $w = \arg\min_{\bar{w}: \mathcal{H}\rightarrow \mathcal{A}} R^{e}(\bar{w} \circ \Phi)$, for all $e\in \mathcal{E}$.
\end{definition}

Given data from training environments $\mathcal{E}_{train}$, the IRM objective aims to find a representation $\Phi$ such that there exists a classifier $w$ that is optimal across all training domains:  
\begin{align}
    \min_{\Phi: \mathcal{X}\rightarrow \mathcal{H},  w:\mathcal{H}\rightarrow \mathcal{Y}} \sum_{e\in \mathcal{E}_{train}} R^{e}(\bar{w} \circ \Phi) & \text{ subject to } w \in \arg\min_{\bar{w}: \mathcal{H}\rightarrow \mathcal{A}} R^{e}(\bar{w} \circ \Phi), \forall e\in \mathcal{E}_{train}  
\end{align}

This represents a challenging, bi-level optimization and \cite{arjovsky2019invariant} propose the IRM-v1 objective which is a practical version to optimize:
\begin{equation}
    \min_{\phi: \mathcal{X} \rightarrow \mathcal{H}} \sum_{e\in \mathcal{E}_{train}}R^{e}(\Phi) + \lambda_{penalty} \cdot \|\nabla_{w|w=1.0} R^{e} (w\cdot \Phi)\|
\end{equation}
Through this optimization, the IRM objective should learn a predictor that only uses the causal parents of the target variable and that is thus invariant across environments. In the supervised setting considered by IRM \cite{arjovsky2019invariant}, $R^{e}$ is the risk of the classifier in environment $e$. For classification and regression problems, this can represent for instance the cross-entropy loss. 

\textbf{Imitation learning risk:} We propose extending IRM to the imitation learning setting by using instead an imitation risk $R^{e}$ as described in equation \ref{eq:imitation_risk}. Moreover, in this setting, our aim is to find an invariant policy $\pi$ across the different environments (instead of the invariant classifier $w$). The risk $R^e$ will therefore be specific to the imitation learning algorithm used. For instance, in behaviour cloning (BC) \cite{pomerleau1991efficient}, for categorical actions $R^e_{\text{BC}} = \text{Cross entropy} (\pi(\cdot \mid x_t), a_t)$. Alternatively, ValueDice(VDICE) \cite{kostrikov2019imitation} minimizes the disparity between the occupancy measure of the expert policy vs the imitator policy  $R^e_{\text{VDICE}} = D_{KL}(\rho^e_D || \rho_{\pi})$. We use the IRMv1 objective in conjunction with the following imitation learning algorithms Behaviour Cloning (BC) \cite{pomerleau1991efficient}, Reward-regularized Classification for Apprenticeship Learning (RCAL) \cite{piot2014boosted}, ValueDice(VDICE) \cite{kostrikov2019imitation} and  Energy-based Distribution Matching (EDM) \cite{jarrett2020strictly}.

%\begin{figure}[H]
%    \centering
%	\includegraphics[scale=0.6]{figs/policy.pdf}
%	\caption{Invariant representation learning and policy. We aim to find a representation $\Phi(x) = s$ such that there exists a policy $\pi(a\mid s)$ that matches the expert occupancy measure across domains.}
%	\label{fig:policy}
%\end{figure}

%Possible extension: Put some attention weights to learn the important features. 

%\begin{figure}%
 %   \centering
%    \subfloat[\centering Decomposing observations into state variable and noise variables]{{\includegraphics[width=6cm]{figs/causal_invariance.pdf} }}%
%    \qquad
%    \subfloat[\centering Specific example]{{\includegraphics[width=6cm]{figs/causal_features.pdf} }}%
 %   \caption{Invariant causal imitation learning.}%
   % \label{fig:causal_imitation}%
%\end{figure}

\newpage
\section{Experimental details} \label{apx:experiments_details}

\subsection{Environments details}

We use the following control tasks from OpenAI gym for experiments \cite{brockman2016openai}: Acrobot\cite{geramifard2015rlpy}, Cartpole\cite{barto1983neuronlike}, Lunar Lander \cite{brockman2016openai} and BeamRider \cite{bellemare2013arcade}. For each task, we use pre-trained RL agents from RL Baselines Zoo \cite{rl-zoo} and Stable OpenAI Baselines \cite{stable-baselines} to obtain expert policies. We provide in Table \ref{tab:envs} details about the different environments used including the size of the observation and action space, the agent used as demonstrator, the demonstrator's performance (average return) as well as the performance of an agent randomly selecting actions. The performance on the OpenAI gym tasks is measured in terms of average return of running the agent in the environment. 

Note that the expert uses the original observation space for each task. To create the different environments with spurious correlations used to train and test the imitation learning benchmarks on Acrobot\cite{geramifard2015rlpy}, Cartpole\cite{barto1983neuronlike} and Lunar Lander \cite{brockman2016openai}, we augment the observation space as follows. We add 3 noise variables that are different multiplicative factors of the last 3 variables in the original state space. For Acrobot, these are cosine of the second rotational angle and the two joint angular velocities, for CartPole these are Cart Velocity, Pole Angle, Pole Angular Velocity and for LunarLander these are Lander angular velocity, leg 1 ground contact and leg 2 ground contact. The invariant causal state is represented by the original variables in the state space of each control task. We train the benchmarks on demonstrations from two environments with $1\times$ and $2\times$ multiplicative factors for the spurious correlations and we test on an environments with multiplicative factors sampled from $\mathcal{U}(-1, 1)$.

Alternatively, for BeamRider \cite{bellemare2013arcade}, we create the different environments for training and testing by using different camera angles, similarly to \cite{zhang2020invariant}. More specifically, we use 2 training environments where the game frames are rotated by 10 degrees to the left and to the right respectively and a test environment that does not have any rotation.

For each OpenAI Gym task, and for each benchmark, we obtain datasets with $N_{traj} \in \{1, 5, 10, 15, 20\}$ demonstrated trajectories from each of the two training environments. We evaluate each benchmark in the test environment by computing the average return over 300 episodes roll-outs. We repeat each experiments 10 times, each time sampling different expert demonstrations for training.

In addition, we also use a dataset from the Medical Information Mart for Intensive Care (MIMIC-III) database \cite{johnson2016mimic}. For each patient, we extract 52 clinical covariates including vital signs (e.g. respiratory rate, heart rate, temperature, O2 saturation) and lab test (e.g. glucose, hemoglobin, magnesium, potassium, platelet count, white blood cell count) that are aggregated every hour during their ICU stay. We consider patient trajectories that are up to 24 hours. Moreover, we concatenate the last 4 hours to build the observations received by each imitation learning algorithm. The expert in this case is the doctor and we consider as action the ventilator support. We consider three environment each with 2000 independent patient trajectories from MIMIC III. Two of the environments are used for training and one for testing. We augment the original feature space by adding 20 spurious correlations (noise variables) that are the same as the expert actions with probabilities $p=0.1$ and $p=0.2$ in the training environments and with probability $p=0.8$ in the testing environment.

%We then follow an approach295similar to the one in [7] to obtain datasets with demonstrations from the expert in two different296environments. In particular, we add spurious correlations to the state space of each control task and297an environment identifier. The spurious correlations in each environment are different multiplicative298factors of a subset of variables in the original state space. The invariant causal state is represented299by  the  original  variables  in  the  state  space  of  each  control  task.   We  train  the  benchmarks  on300demonstrations from two environments with1×and2×multiplicative factors for the spurious301correlations and we test on an environments with multiplicative factors sampled fromU(−1,1).302Further details about the train and test environments can be found in Appendix E.

%==============================================================================
\begin{table}[t]\small
%==============================================================================
\newcolumntype{H}{>{          \arraybackslash}m{2.4 cm}}
\newcolumntype{I}{>{\centering\arraybackslash}m{2.8 cm}}
\newcolumntype{J}{>{\centering\arraybackslash}m{2.0 cm}}
\newcolumntype{K}{>{\centering\arraybackslash}m{2.0 cm}}
\newcolumntype{L}{>{\centering\arraybackslash}m{2.3 cm}}
\newcolumntype{M}{>{\centering\arraybackslash}m{2.4 cm}}
\setlength\tabcolsep{1.35pt}
\renewcommand{\arraystretch}{1.02}
\begin{adjustbox}{max width=\textwidth}
\begin{tabular}{H|IJKLM}
\toprule
\textit{Environments} & {Original Obs. Space} & {Action Space} & {Demonstrator} & {Random Perf.} & {Demonstrator Perf.} \\
\midrule
Acrobot-v1     & Continuous (6)                       & Discrete (3) & PPO2 Agent   & $-439.92\pm13.14$ & $-87.32\pm12.02$   \\
CartPole-v1    & Continuous (4)                       & Discrete (2) & DQN Agent    & $19.12\pm1.76$    & $500.00\pm0.00$    \\
LunarLander-v2 & Continuous (8)                       & Discrete (4) & PPO2 Agent   & $-452.22\pm61.24$ & $271.71\pm17.88$   \\
BeamRider-v4 & Continuous ($210\times 160\times 3$) & Discrete (9) & PPO2 Agent & $754.84 \pm 214.85$ & $1623.80 \pm 482.27$  \\
MIMIC-III   & ~Continuous (208)                     & Discrete (2) & Clinician & -                & -                 \\
\bottomrule
\end{tabular}
\end{adjustbox}
\vspace{0.5em}
\caption{Environment details. The random and demonstrator performances are averaged over 1,000 episodes roll-outs.}
\vspace{-1.0em}
\label{tab:envs}
\end{table}

%==============================================================================
\newpage
\subsection{Implementation details}

Similarly to \cite{jarrett2020strictly} and for a fair comparison, whenever possible, we use the same policy network architecture for all imitation learning benchmarks. 
For Acrobot\cite{geramifard2015rlpy}, Cartpole\cite{barto1983neuronlike}, Lunar Lander \cite{brockman2016openai} and MIMIC-III \cite{johnson2016mimic} we use a policy network consisting of two fully-connected hidden layers with 64 units each and with ELU activation. Alternatively, for BeamRider \cite{bellemare2013arcade}, we use as the policy network a convolutional neural network with 3 convolutional layers consisting of 32-64-64 filters, followed by a fully connected layer of size 64, with all layers followed by ReLU activations.

We consider discrete actions in all environments; thus, the output layer of the policy network has the same number of dimensions as the action space.  For all the different environments used for evaluation we optimize the parameters using the Adam Optimizer for 10k iterations with learning rate $\lambda = 0.001$ and batch size 64 \cite{jarrett2020strictly}. Moreover, we use the publicly available code for the different benchmarks used and other than the standardized policy network, we keep the optimal hyperparameters in the original implementations.

The experiments were run on a system with 6CPUs, an Nvidia K80 Tesla GPU and 56GB of RAM.

\textbf{Invariant Causal Imitation Learning:} uses a policy network as described above and neural network architectures with two fully connected hidden layers with 64 units and with ELU activation for each of $\phi$, $\mu^e$, $g_s$, $g_{\eta}$, $\psi$, $c_s$ and $T_{\theta_m}$ in the Acrobot\cite{geramifard2015rlpy}, Cartpole\cite{barto1983neuronlike}, Lunar Lander \cite{brockman2016openai} and MIMIC-III \cite{johnson2016mimic} experiments. On the other hand, for the BeamRider environment \cite{bellemare2013arcade}, we use a convolutional neural network with 3 convolutional layers consisting of 32-64-64 filters, followed by a fully connected layer of size 64, with all layers followed by ReLU activations for $\phi$ and $\mu^e$. Moreover, we use neural network architectures with two fully connected hidden layers with 64 units and with ELU activation for $g_s$, $g_{\eta}$, $c_s$, $T_{\theta_m}$ and for the policy network $\pi$. Finally, for $\psi$ we use a neural network with 2 fully connected layers of 64 and $64 \times 7 \times 7$ hidden units followed by  3 transposed convolution layers consisting of 64-64-32 filters.

\textbf{Behaviour cloning (BC):} We implement behaviour cloning by using a policy network architecture as described above. We train the model using cross entropy loss and we optimize it as described above. We use the same hyperparameters for BC-IRM but instead use IRM-v1 objective. 

\textbf{Reward-regularized Classification for Apprenticeship Learning (RCAL):} We implement RCAL by adding a sparsity-based loss on the implied rewards \cite{piot2014boosted} and we set the sparsity-based regularization coefficient to 0.01. We use the same policy network architecture for optimization procedure as described above. Moreover, we use the same hyperparameters for RCAL-IRM but instead use IRM-v1 objective.

\textbf{Energy-based Distribution Matching (EDM):} We use the publicly available implementation for EDM \cite{jarrett2020inverse} from here: \url{https://github.com/vanderschaarlab/mlforhealthlabpub}. We use the same policy network and optimization procedure as above, which corresponds to the ones used in \cite{jarrett2020inverse}. Moreover, following the implementation details provided in \cite{jarrett2020inverse} we set the joint EBM training hyperparameters to noise coefficient $\sigma=0.01$, buffer size $\kappa = 10000$, length $l=20$, re-initialization $\delta=0.05$ and SGLD step size $\alpha=0.01$. For EDM-IRM we use the same hyperparameters, but instead optimize the IRM-v1 objective.  

\textbf{ValueDice (VDICE):} We use the publicly available implementation for VDICE \cite{kostrikov2019imitation} from \url{https://github.com/google-research/google-research/tree/master/value_dice}. However, to adapt the model to discrete actions we modify the last layer of the actor network to use Gumbel-softmax. For Acrobot\cite{geramifard2015rlpy}, Cartpole\cite{barto1983neuronlike}, Lunar Lander \cite{brockman2016openai} and MIMIC-III \cite{johnson2016mimic}, the actor and discriminator network architecture used have two fully connected hidden layers with 64 units and ReLU activation. Conversely, for BeamRider \cite{bellemare2013arcade} the actor and discriminator networks consist of 3 convolutional layers consisting of 32-64-64 filters, followed by a fully connected layer of size 64, with all layers followed by ReLU activations. As described in \cite{kostrikov2019imitation}, orthogonal regularization is used for the actor and a learning rate of 0.00001. The discriminator uses a learning rate of 0.001. For VDICE-IRM we use the same hyperparameters, but instead optimize the IRM-v1 objective.

\newpage
\section{Additional experiments} \label{apx:additional_experiments}

\subsection{Ablations}

To understand the impact of the different components in the overall loss function used to train ICIL, we performed an ablation experiment on the CartPole \cite{barto1983neuronlike} control task from OpenAI gym \cite{brockman2016openai}. We follow the same training and testing set-up described in Section \ref{sec:eval_openai_gym}. 

\begin{figure}[h]
		\begin{center}
			\includegraphics[width=0.6\columnwidth]{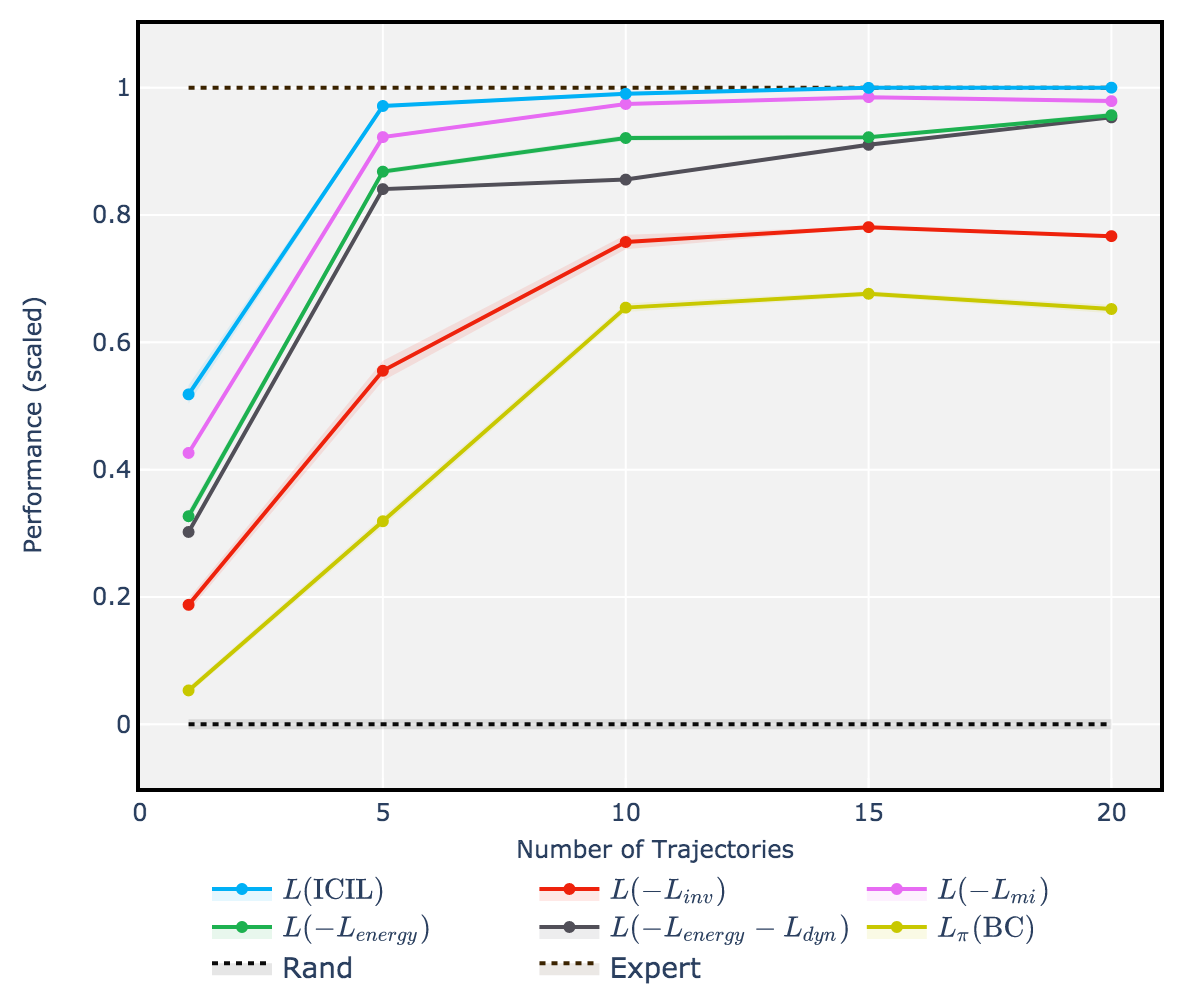}
		\end{center}
		\caption{Evaluating the impact of the different loss components on overall performance for the CartPole control task. $x$-axis indicates the number of trajectories (in $\{1, 5, 10, 15, 20\}$) with expert demonstrations from each training environment given as input to each ablated version of ICIL and $y$-axis represents average return of the learnt imitation policy on the test environments, scaled between 1 (expert performance) and 0 (random policy performance).} 
		\vspace{-0.4cm} 
		\label{fig:results_ablations}
\end{figure}

Let $L = L_{\pi} + L_{\text{dyn}} + L_{\text{inv}} + L_{\text{mi}} + L_{\text{energy}}$ be the full loss function used for training ICIL. Refer to Section \ref{sec:learning_invariant_causal_rep} for details of how each component in $L$ is defined. The results in Figure \ref{fig:results_ablations} illustrate the impact of removing different terms from this loss function on overall performance (average return of the learnt imitation policy in the test environment). The average return is scaled between 1 (expert performance) and 0 (random policy performance). The setting of only using $L_{\pi}$ corresponds to the Behaviour Cloning (BC) benchmark. 

We notice that while each term in the loss $L$ used to train ICIL is important for the overall performance, the loss term $L_{inv}$ which ensures that the state representation  is invariant across environments plays the most significant role on the performance in the test environment. This is due to the fact that $L_{inv}$ is crucial for learning the shared latent structure across the different environments that consists of the causal parents of the actions. 

\subsection{Train vs. test performance}

\begin{figure}[t]
    \centering
	\subfloat[ICIL]{\includegraphics[width=5cm]{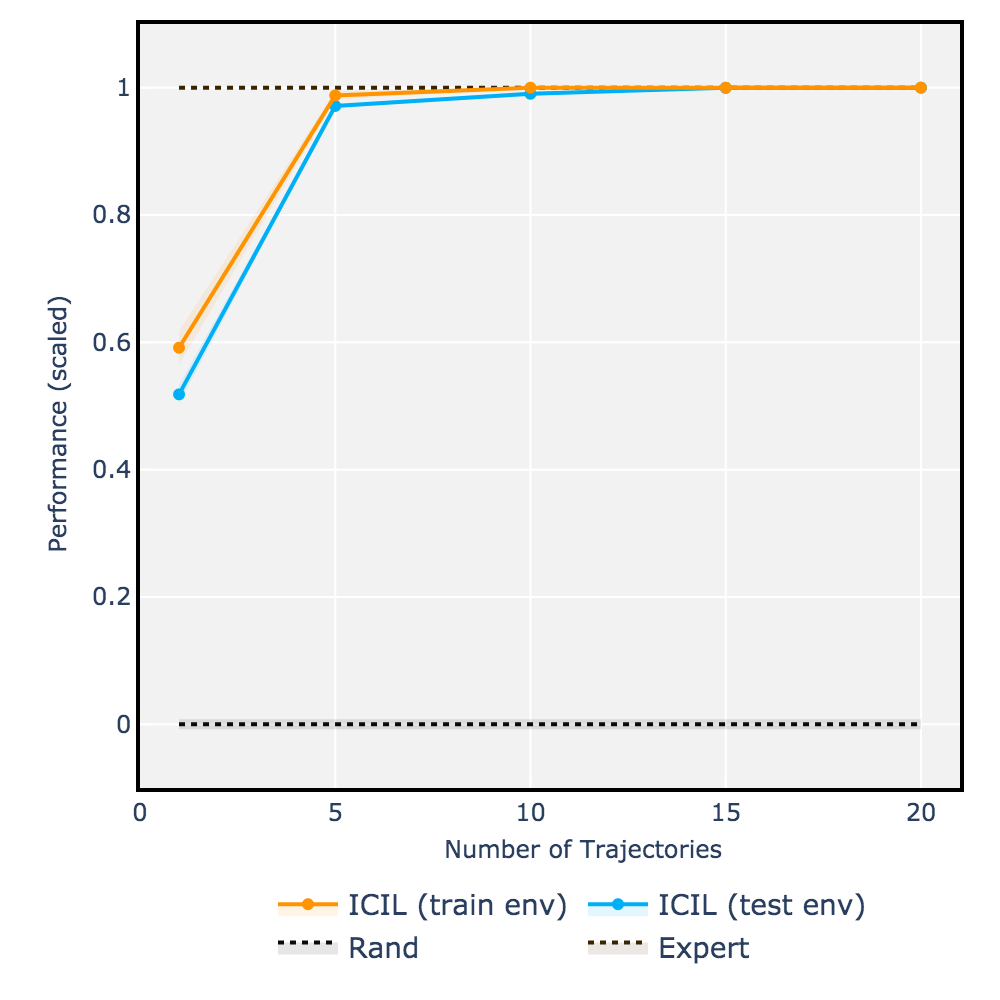}}%
	\subfloat[BC]{\includegraphics[width=5cm]{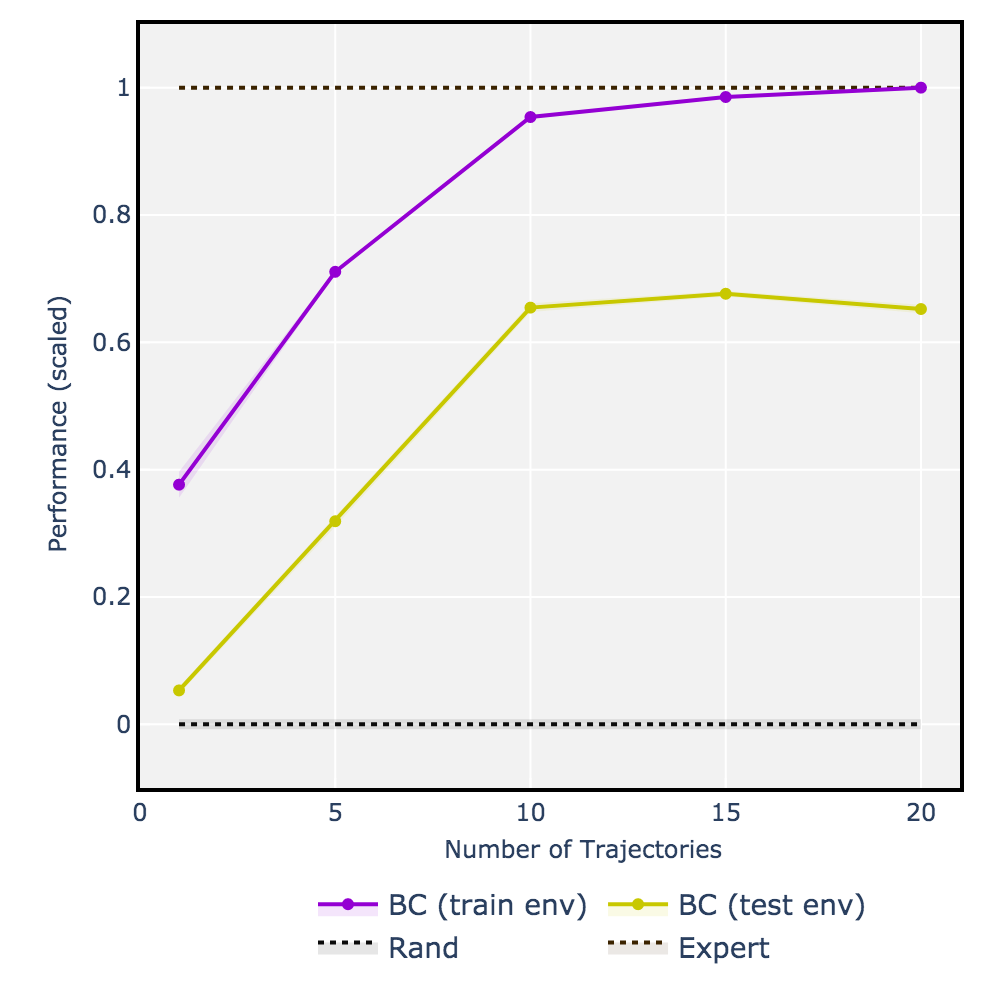}}%
	\caption{Train vs. test performance on CartPole. $x$-axis indicates the number of trajectories (in $\{1, 5, 10, 15, 20\}$) with expert demonstrations from each training environment given as input to each benchmark and $y$-axis represents average return of the learnt imitation policy when evaluated on both the train and test environments, scaled between 1 (expert performance) and 0 (random policy performance). }    \label{fig:results_train_test}%
	\vspace{-4mm}
\end{figure}

We report here the evaluation metrics for both training and testing environments on the CartPole \cite{barto1983neuronlike} control task. We follow the same experimental set-up described in Section \ref{sec:eval_openai_gym}. In Figure \ref{fig:results_train_test} we report the performance of ICIL and BC  when evaluated both on 300 new episode roll-outs from one of the environments used for training as well as when evaluated on a test environment with different multiplicative factors for the noise variables. We notice that the imitation policy learnt by BC relies on the spurious correlations and thus fails to generalize beyond the environments it has been exposed to. On the other hand, ICIL learns an imitation policy that correctly depends on the state variables, which are the ones that are being shared across the different environments. 

\newpage

\subsection{Robustness to increasing the size of spurious correlations}

\begin{figure}[h]
		\begin{center}
			\includegraphics[width=0.5\columnwidth]{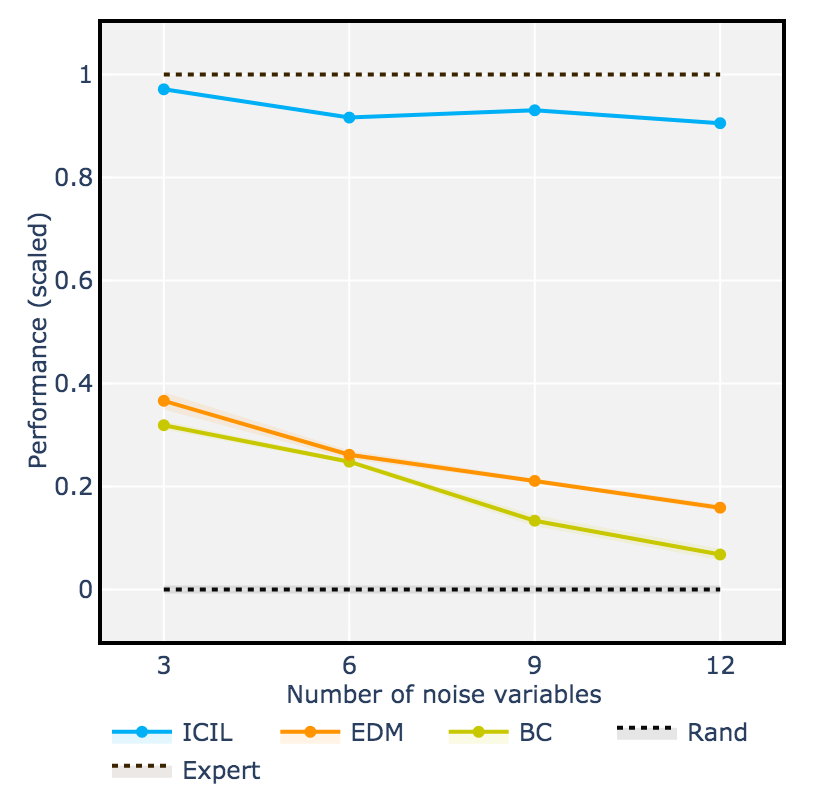}
		\end{center}
		\caption{Robustness to increasing the number of noise variables. $x$-axis indicates the number of noise variables (in $\{3, 6, 9, 12\}$) that are part of the observations in each environment and $y$-axis represents average return of the learnt imitation policy on the test environments, scaled between 1 (expert performance) and 0 (random policy performance).  } 
		\vspace{-0.4cm} 
		\label{fig:results_noise}
\end{figure}

Finally, we investigate the robustness of the different imitation learning methods to increasing the size of the spurious correlations (i.e. the number of noise variables in each environment). In this case, we consider again the CartPole \cite{barto1983neuronlike} control task and the setting where 5 trajectories from each training environment are given as input to each benchmark during training. The spurious correlations in each environment are different multiplicative factors of the last 3 variables in the original state space of the control task. We follow the same set-up described in Appendix \ref{apx:experiments_details} for setting the multiplicative factors for the train and test environments. Figure \ref{fig:results_noise} illustrates the performance of ICIL, BC and EDM when increasing the number of noise variables used for the different environments. We notice that ICIL is robust to having more spurious correlations, while the performance of BC and EDM degrades more significantly in the case where the observations from the expert demonstrations have a large number of noise variables. 

\end{document}